\documentclass[letterpaper,journal]{IEEEtran}
\usepackage{amsmath,amsfonts}
\usepackage{algorithmic}
\usepackage{algorithm}
\usepackage{array}
\usepackage[caption=false,font=normalsize,labelfont=sf,textfont=sf]{subfig}
\usepackage{textcomp}
\usepackage{stfloats}
\usepackage{url}
\usepackage{verbatim}
\usepackage{graphicx}
\usepackage{cite}

\usepackage{xcolor}
\usepackage[table]{colortbl}
\usepackage{amssymb}
\usepackage{hyperref}
\usepackage{hhline}
\usepackage{graphicx}
\usepackage{multirow}
\usepackage{longtable}
\usepackage{pifont}
\usepackage[T1]{fontenc}

\definecolor{data-1}{rgb}{0,0,1}

\newcommand{\MM}[1]{\textcolor[rgb]{0,0,0}{#1}}

\begin{document}

\title{\MM{A Comprehensive Taxonomy and Analysis of Talking Head Synthesis: Techniques for Portrait Generation, Driving Mechanisms, and Editing}}

\author{Ming Meng, Yufei Zhao, Bo Zhang, Yonggui Zhu, Weimin Shi, Maxwell Wen, and Zhaoxin Fan$^*$\thanks{$*$ Corresponding author.}
\thanks{This paper was produced by the IEEE Publication Technology Group. They are in Piscataway, NJ.}

}

\markboth{Journal of \LaTeX\ Class Files,~Vol.~14, No.~8, June~2024}%
{Shell \MakeLowercase{\textit{et al.}}: A Sample Article Using IEEEtran.cls for IEEE Journals}


\maketitle

\begin{abstract}
Talking head synthesis, an advanced method for generating portrait videos from a still image driven by specific content, has garnered widespread attention in virtual reality, augmented reality and game production. Recently, significant breakthroughs have been made with the introduction of novel models such as the transformer and the diffusion model. Current methods can not only generate new content but also edit the generated material. This survey systematically reviews the technology, categorizing it into three pivotal domains: portrait generation, driven mechanisms, and editing techniques. We summarize milestone studies and critically analyze their innovations and shortcomings within each domain. Additionally, we organize an extensive collection of datasets and provide a thorough performance analysis of current methodologies based on various evaluation metrics, aiming to furnish a clear framework and robust data support for future research. Finally, we explore application scenarios of talking head synthesis, illustrate them with specific cases, and examine potential future directions. 
\end{abstract}


\section{Introduction}
\label{sec:introduction}
The rapid advancements in deep learning have catalyzed significant progress in the field of computer vision. One of the landmark innovations in this domain is the development of Generative Adversarial Networks (GANs) \cite{goodfellow2020GAN}, which has dramatically transformed image generation and manipulation tasks. This technology extends beyond simple image editing techniques, such as style transfer, image inpainting, and super-resolution, to the sophisticated generation of entirely new images from textual descriptions or random noise. Among these generative tasks, one of the most compelling challenges is the creation of realistic and natural human portraits that avoid the uncanny valley effect. Talking head synthesis, a specialized application within this sphere, focuses on generating dynamic, speaking portraits from static images, and has found widespread application in areas such as film production, gaming industry, and virtual hosting.

Talking head synthesis encompasses three primary stages: portrait generation, driving mechanisms, and editing techniques. The initial stage, portrait generation, involves creating new facial images using advanced generative models such as GANs, Variational Autoencoders (VAEs) \cite{kingma2013auto}, and Diffusion Models \cite{ho2020denoising}. These models enable the production of high-quality static images that serve as the foundation for subsequent animation. The second stage, driving mechanisms, is critical for animating these static images. This process can be driven by video or audio inputs, which dictate the movements and expressions of the generated portraits. Video-driven approaches replicate the facial movements and expressions of a character from a driving video \cite{thies2019deferred,wang2018video2video}, while audio-driven methods synchronize lip movements and facial expressions with an audio track \cite{li2021write,yi2020audio}. Text inputs, often converted to audio before driving, also play a significant role in this stage. Key factors in this process include preserving the identity of the source image, accurately replicating head movements, and capturing the nuances of facial expressions. The final stage, editing techniques, allows for the modification of generated portraits to enhance realism and suitability for specific applications. These techniques enable the adjustment of attributes such as emotions, head orientation, and eye blinking without necessitating retraining of the model \cite{lee2023lpmm,tang20233dfaceshop,cheng2022videoretalking,sung2024laughtalk}. This flexibility is crucial for refining the results and achieving high-quality outputs.

Despite the remarkable advancements in talking head synthesis, several challenges persist. These include maintaining temporal consistency, handling large-angle head poses, and addressing data biases. Furthermore, the reliance on large pretrained models poses limitations in terms of adaptability and computational requirements. Addressing these challenges is essential for advancing the field and expanding its applications. In this comprehensive survey, we systematically review and categorize the current state of talking head synthesis technology. We delve into the three pivotal domains of portrait generation, driving mechanisms, and editing techniques, highlighting key studies and critically analyzing their innovations and limitations. Additionally, we compile and organize extensive datasets, providing a thorough performance analysis based on various evaluation metrics. We also explore the diverse application scenarios of talking head synthesis, presenting specific case studies and discussing future research directions. This survey aims to provide a clear framework and robust data support for researchers and practitioners, facilitating further advancements in this rapidly evolving field.

\begin{figure*}[!ht]
\centering
\includegraphics[width=1.0\textwidth]{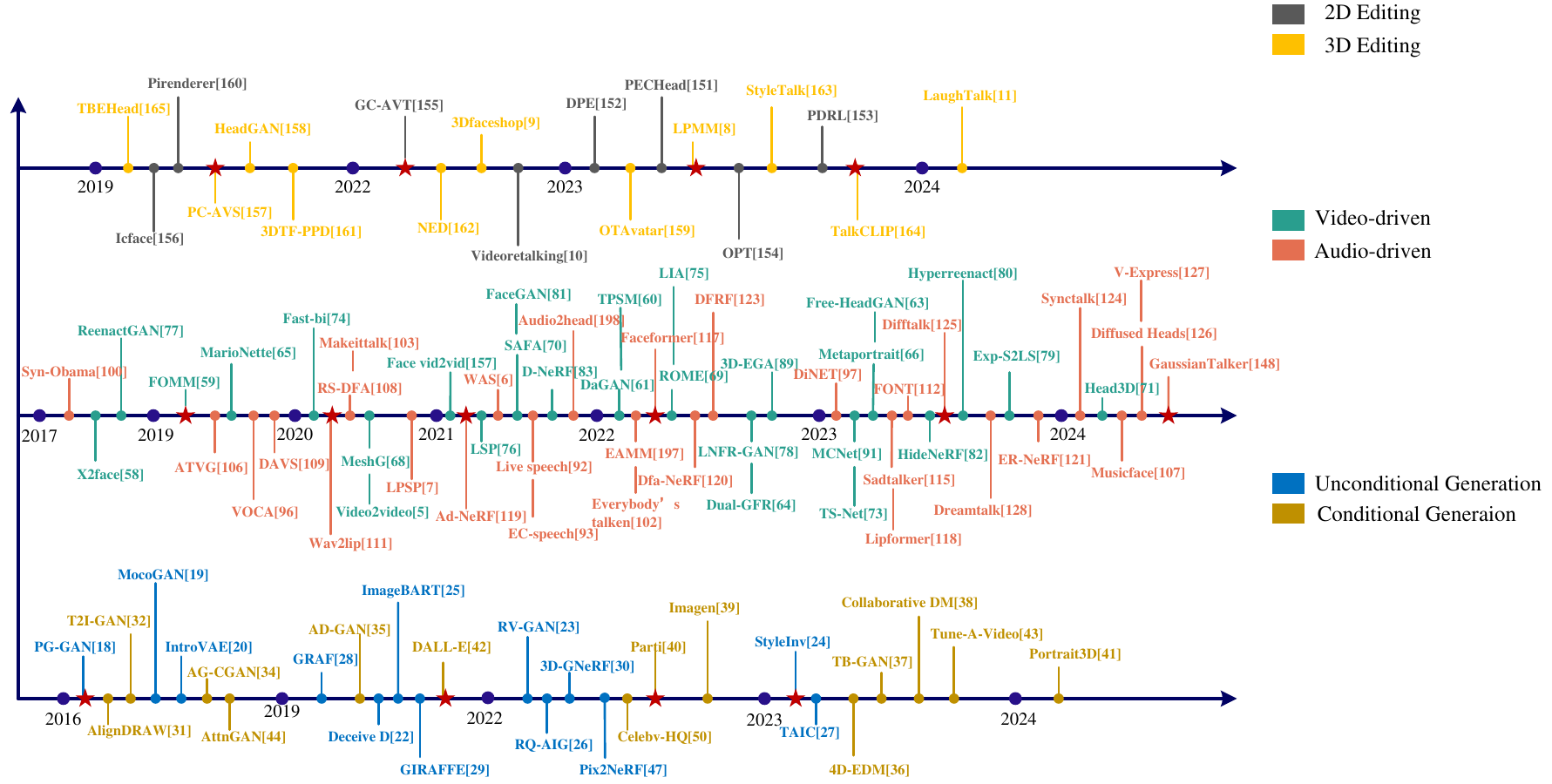}
\caption{Chronological overview of the talking head synthesis's three primary stages. The bottom-up order is portrait generation, driven mechanisms and editing techniques. Pentagrams mark more influential methods in each stage.}
\label{fig:timeline}
\end{figure*}

\subsection{Comparison with Existing Surveys}
\label{subsec:Survey comparison}
\MM{In this section, we provide a comparative analysis of our work with recent surveys on talking head synthesis.}

\MM{Kammoun et al. \cite{kammoun2022generative} reviewed the application of GANs in face generation, focusing on direct face synthesis using GANs. In contrast, our survey delves into the utilization of audio and video inputs for facial animation, offering a broader perspective.}

\MM{Zhang et al. \cite{zhang2023text} examined text-guided image generation with diffusion models, highlighting the advancements in generating images from textual descriptions. Our work, however, centers on the dynamic aspect of facial animation driven by audio and video, presenting a different angle of the generative model applications.}

 \MM{Liu et al. \cite{liu2023audio} provided a comprehensive overview of audio-driven talking head synthesis. While our paper acknowledges the contributions in audio-driven methods, it expands the scope to include video-driven techniques, offering a more holistic view of the field.}

 \MM{Sha et al. \cite{sha2023deep} and Gowda et al. \cite{gowda2023pixels} contributed extensive reviews on talking head synthesis. Our paper advances the discourse by incorporating a detailed examination of editing techniques and the latest methodologies for integrating diffusion models into talking head synthesis.}

\MM{Our review offers a systematic exploration of the three pivotal components of talking head synthesis: portrait generation, driving mechanisms, and editing techniques. We meticulously analyze the application of avant-garde technologies, such as GANs and diffusion models, to assess their current capabilities and delineate potential applications. Moreover, we underscore the existing challenges and chart a course for future research directions.}
\subsection{Organization of This Article}
\label{subsec:organization of this paper}
\begin{figure*}[t]
    \centering
    \includegraphics[width=1.0\textwidth]{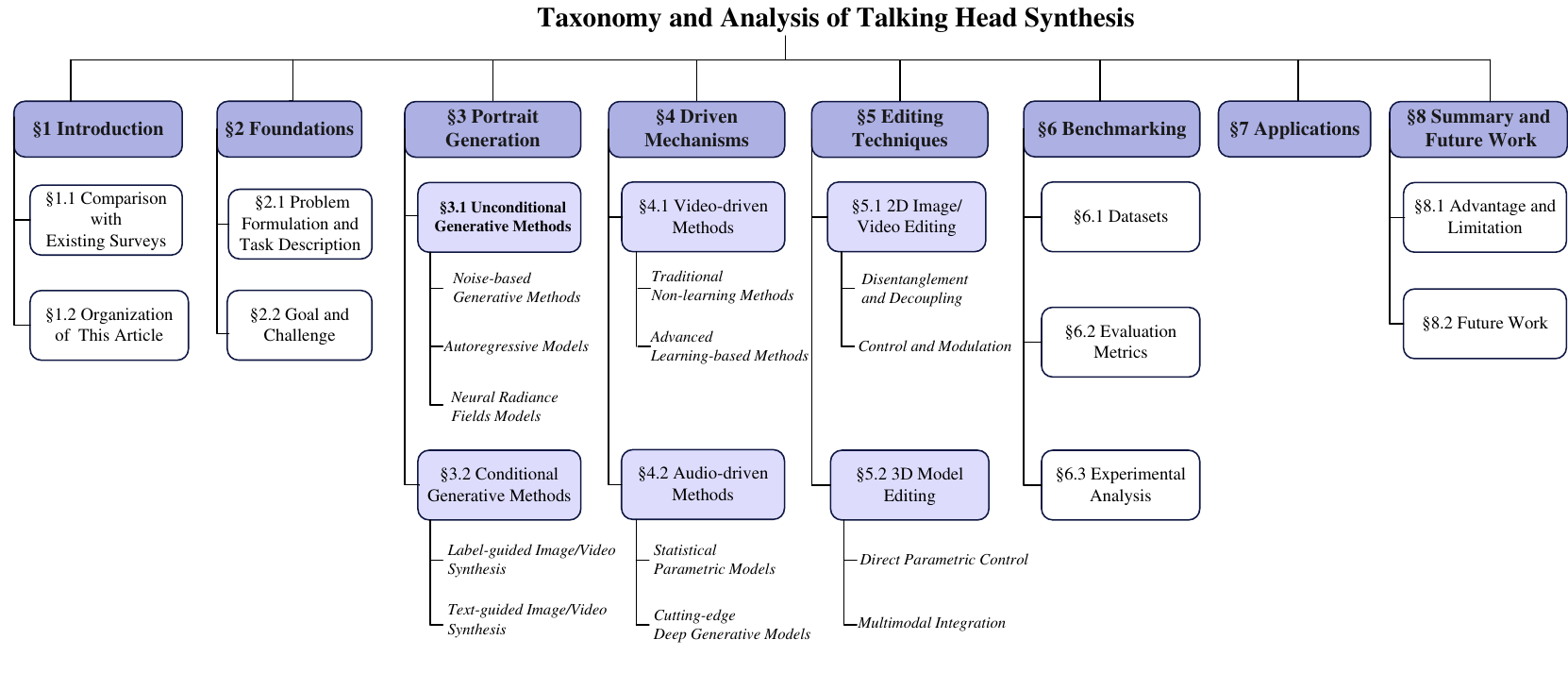}
    \caption{An overview of this survey.}
    \label{fig: Organization}
\end{figure*}

\MM{Our paper presents a comprehensive survey of talking head synthesis. Section \ref{sec:foundations} introduces the foundations, covering fundamental algorithmic processes and task descriptions (\ref{subsec:problem formulation}) and summarizing existing problems and challenges (Section \ref{subsec:challenge}). Section \ref{sec: portrait generation} explores generative methods, detailing unconditional methods (Section \ref{subsec: unconditional generative methods}) and conditional methods (Section \ref{subsec: conditional generative methods}). Section \ref{sec: Driven Mechanisms} describes animation driving methods, including video-driven (Section \ref{subsec:video-driven}) and audio-driven techniques (Section \ref{subsec:audio-driven}). Section \ref{sec:editing techniques} reviews editing techniques, divided into 2D Image/Video Editing (Section \ref{subsec: 2D editing}) and 3D Model Editing (Section \ref{subsec:3D editing}). Section \ref{sec:benchmarking} presents benchmarks, including datasets, evaluation metrics, and experimental analysis. Section \ref{sec:application} discusses applications, while Section \ref{sec: summary and future work} addresses the advantages, limitations, existing issues, and potential solutions for future work. Finally, Section \ref{sec:conclusion} concludes the survey with conclusive remarks, encapsulating the collective findings and underscoring the significance of the research in the domain of talking head synthesis. This organization is visually represented in Figure \ref{fig: Organization}.}

\section{Foundations}
\label{sec:foundations}
\subsection{Problem Formulation and Task Description}
\label{subsec:problem formulation}

\begin{figure*}[!ht]
    \centering
    \includegraphics[width=1.0\textwidth]{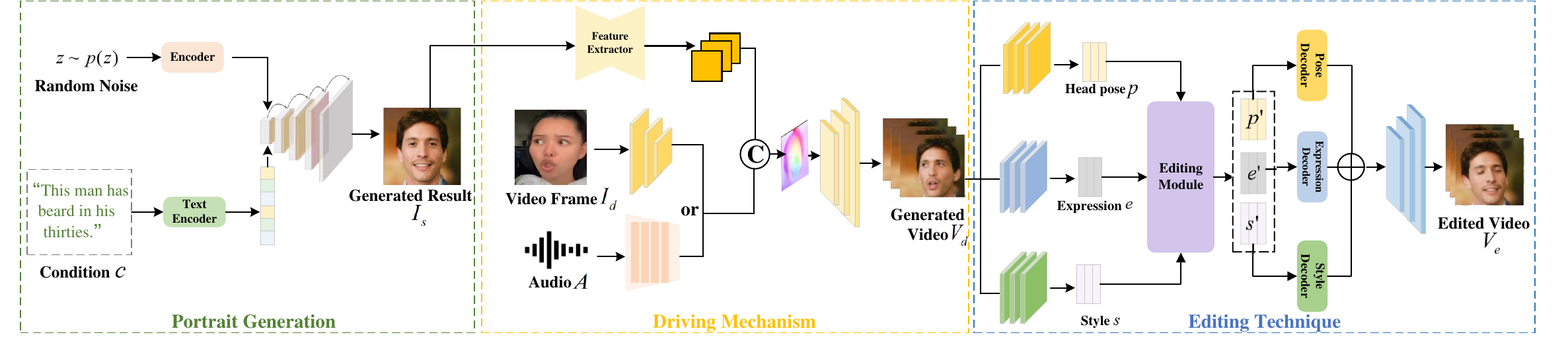}
    \caption{\MM{The comprehensive process of talking head synthesis involves three key stages. First, portrait generation creates a static image from random noise, optionally incorporating specified attributes. Second, the driving mechanism animates this image using video frames or audio content through the extraction and application of intermediate representations like keypoints and landmarks. Third, the editing technique refines the animated output, enhancing visual coherence and quality by precisely adjusting pose and expression parameters.}}
    \label{fig:Structure}
\end{figure*}

\MM{Talking head synthesis is a multifaceted task designed to generate, animate, and refine realistic human head images based on specific conditions and driving content. This process is crucial for applications such as virtual avatars, video conferencing, and entertainment. The pipeline can be divided into three primary stages: talking head portrait generation, talking head driving, and talking head editing. Each stage plays an essential role in ensuring that the final output is both visually coherent and dynamically expressive. The entire process of talking head synthesis is illustrated in Figure \ref{fig:Structure}.}
\MM{In this context, we use a generative model to create a portrait of a person from random noise $z \sim p(z)$. When specific results are desired, additional conditions can be included as inputs to the model. The image generation process can be described as follows:}
\begin{equation}
    I_s=G_i(z,c;\beta_{G_{i}})
\end{equation}
where $I_s$ is the generated image, $c$ is the condition as a guidance for generation, $G_i$ is the generative model and $\beta_{G_{i}}$ means model's parameters.

In the process of talking head driving, there are mainly two driven contents: video and audio.
\MM{For video-driven methods, both a source image and a driving video $V_d$ are required. We select the generated image $I_s$ as the source image, although any portrait image can be used. The frames in the video are denoted as $I_d$. Using $I_s$ and $I_d$, we aim to generate a new image $I_g$. The generation process can be divided into four detailed steps: intermediate representation, feature extraction, transformation, and image generation. Firstly, the source image $I_s$ and the video frame $I_d$ are processed by a network $F$, which predicts or detects intermediate representations, such as keypoints, landmarks, and 3D meshes. This can be formulated as:}
\begin{equation}
    R_s=F(I_s;\beta_F),  R_d=F(I_d;\beta_F)
\end{equation}
where $R_s$ and $R_d$ are the intermediate representations of the source image and video frame, respectively, $\beta_F$ is the parameters of the network $F$. Then, we will extract important features from driving frame's intermediate representation, such as head movements and facial expression. This can be expressed as:
\begin{equation}
    f_d=P(R_d;\beta_P)
\end{equation}
where $f_d$ is the features of $R_d$, $P$ is the extracting network and $\beta_P$ is the parameters of it. The extracted features will be used to transform the intermediate representation of the source image, the new intermediate representation $\overline{R}_s$ preserves the identity of the source image but has same pose and expression with the video frame. This can be described as:
\begin{equation}
    \overline{R}_s=T(R_s,f_d;\beta_T)
\end{equation}
where $T$ is the transformation network and $\beta_T$ is the parameters of it. Finally, we need convert the intermediate representation $\overline{R}_s$ to the image $I_g$ and obtain the generated $V_d$. This can be expressed as:
\begin{equation}
    I_g=G(\overline{R}_s;\beta_G)
\end{equation}
where $G$ is the generative network and $\beta_G$ is the parameters of it. For audio-driven talking head synthesis, the input is a source image $I_s$ and a audio clip $A$ used for driving. The processing of the source image is the same as in video-driven methods, first obtaining an intermediate representation, then it is deformed under the influence of features extracted from the audio. However, the processing of audio is more complex as audio and image belong to different modalities of information. It is necessary to firstly extract features from the audio, such as the Mel Spectrogram, Fundamental Frequency (F0), and vocal intensity. Afterwards, a mapping model is needed to convert the extracted audio features into control signals for facial movements. These steps can be represented as:
\begin{equation}
    \overline{R}_s=T(R_s,C(M(A)))
\end{equation}
where $M$ is an audio feature extractor and $C$ is a mapping network. Then $\overline{R}_s$ will be the input to the generative network to get $I_g$. Overall, audio-driven talking head synthesis focuses on utilizing audio features to generate accurate mouth movements and corresponding facial expressions, while video-driven methods emphasize capturing more comprehensive facial movements and expressions from video.

For the task of editing, we can understand it as a higher-level operation. Not only can it accomplish driving, but it also adds an editing module. The module enables us to modify specific attributes during talking head synthesis. Typically, editing requires the control of decoupled parameters to generate an newly edited video $V_e$, such as the parameters of 3D morphable model (3DMM) \cite{blanz1999morphable}. This task can be clearly described as: 
\begin{equation}
    I_e=E(I_g;\beta_E)
\end{equation}
where $E$ is the editing module, $I_e$ is the edited image and $\beta_E$ is the parameters of it.

\subsection{Goal and Challenge}
\label{subsec:challenge}
In this section, we delve into the core objectives and key challenges faced in the field of talking head synthesis. Initially, the primary goal is to achieve high realism, ensuring that the generated images or videos are nearly indistinguishable from real world. Additionally, these avatars should display a rich array of expression and movements, capturing subtle emotional nuances to offer a more authentic and expressive visual experience. The need for real-time interaction is increasing quickly, new methods should ensure that the avatars can respond instantaneously to external inputs, facilitating seamless interaction. Personalization is also crucial, allowing the avatars to be tailored according to specific user requirements, digital human is a typical application  of it.

Conversely, achieving these goals is fraught with challenges. For deep learning training, the diversity and quality of data pose significant hurdles, high-quality and varied training data are fundamental to enhancing realism. Moreover, the demand for computational resources is often substantial, representing a significant barrier for general application. Because of different modalities, ensuring perfect synchronization between the audio and facial movements presents a technical challenge, as any minor discrepancies can severely impact high fidelity. Overcoming these challenges will broaden the application prospects and enhance user satisfaction in talking head synthesis technology.

\section{Portrait Generation}
\label{sec: portrait generation}
\MM{Portrait generation focuses on the automated creation of digital images depicting human faces, primarily utilizing generative models, with GANs being the most prominent. Research in this field is divided into two main categories based on the nature of the input content: unconditional generative methods and conditional generative methods.} 
Table \ref{table: portrait generation} summarizes the representative works on portrait generation.
\subsection{Unconditional Generative Methods}
\label{subsec: unconditional generative methods}
\MM{Unconditional generative methods create digital portraits without relying on specific input conditions such as attributes, labels, or guiding text. These methods leverage random noise or inherent model structures and can be grouped into three main categories: noise-based generative methods, which utilize random noise; autoregressive models, which model probability distributions; and NeRF models, which represent scenes as continuous volumetric functions.} 

\begin{table*}[t]
\centering
\caption{The characteristics of representative portrait generation works, where the Models, Output's Resolution (Outputs.R), Output's Type (Outputs.T) and Controllable, which means controlling certain attributes during generation.}
\label{table: portrait generation}
\scalebox{0.8}{
\resizebox{\textwidth}{!}{%
\begin{tabular}{c|c|c|c|c}
\hline

\multicolumn{1}{c|}{\rule{0pt}{1.5em}\raisebox{0.3em} {Method}} & 
\multicolumn{1}{c|}{\rule{0pt}{1.5em}\raisebox{0.3em} {Models}} & 
\multicolumn{1}{c|}{\rule{0pt}{1.5em}\raisebox{0.3em} {Outputs.R}} & 
\multicolumn{1}{c|}{\rule{0pt}{1.5em}\raisebox{0.3em}{Outputs.T}} & 
\multicolumn{1}{c}{\rule{0pt}{1.5em}\raisebox{0.3em} {Controllable}} \\ 
\hline 

\multicolumn{5}{c}{\textbf{Unconditional Generative Methods}} \\ \hline

PG-GAN.\cite{karras2017progressive} & GAN & 1024×1024 & Image &    \\ \hline
MoCoGAN.\cite{tulyakov2018mocogan} & GAN & N/A & Video & \checkmark   \\ \hline
IntroVAE.\cite{huang2018introvae} & VAE & High-resolution & Video &    \\ \hline
StyleGAN.\cite{karras2020analyzing} & GAN & N/A & Image & \checkmark  \\ \hline
Deceive D.\cite{jiang2021deceive} & GAN & 256×256 & Image &    \\ \hline
RV-GAN.\cite{gupta2022rv}& GAN & N/A & Video &   \\ \hline
StyleInV.\cite{wang2023styleinv} & Diffusion & High-resolution & Image &   \\ \hline
ImageBART.\cite{esser2021imagebart} & Autoregressive & 256×256 & Image & \checkmark   \\ \hline
RQ-AIG.\cite{lee2022autoregressive}& Autoregressive+VAE & 256×256 & Image &    \\ \hline
TAIC.\cite{huang2023towards} & Autoregressive & N/A & Image &    \\ \hline
GRAF.\cite{schwarz2020graf} & GAN & High-resolution & Image & \checkmark   \\ \hline
GIRAFFE.\cite{niemeyer2021giraffe} & GAN & High-resolution & Image & \checkmark  \\ \hline
3D-GNeRF.\cite{bergman2022generative} & GAN & N/A & Image &    \\ \hline

\multicolumn{5}{c}{\textbf{Conditional Generative Methods}} \\ \hline

AlignDRAW\cite{mansimov2015generating} & RNN & N/A  & Image & \checkmark  \\ \hline
T2I-GAN.\cite{reed2016generative} & GAN & N/A & Image & \checkmark  \\ \hline
Composer.\cite{huang2023composer} & Diffusion & 1024×1024 & Image & \checkmark  \\ \hline
AG-CGAN.\cite{bansal2018recycleGAN} & GAN & High-resolution & Image & \checkmark  \\ \hline
AD-GAN.\cite{men2020controllable} & GAN & High-resolution & Image & \checkmark  \\ \hline
4D-EDM.\cite{zou20234d} & Diffusion & High-resolution & Image & \checkmark \\ \hline
TB-GAN.\cite{canfes2023text}& GAN & High-resolution & Image & \checkmark  \\ \hline
Collaborative-DM.\cite{huang2023collaborative} & Diffusion & 512×512 & Image & \checkmark  \\ \hline
Imagen.\cite{ho2022imagen} & Diffusion & 1280×768 & Iideo & \checkmark  \\ \hline
Parti.\cite{yu2022scaling} & Autoregressive & 1024×1024 & Image & \checkmark \\ \hline
Portrait3D.\cite{wu2024portrait3d} & GAN & N/A & Image &   \\ \hline
DALL-E.\cite{ramesh2021zero} & Autoregressive & High-resolution & Image  & \checkmark \\ \hline
Tune-A-Video.\cite{wu2023tune} & Diffusion & N/A & Video & \checkmark  \\ \hline
AttnGAN.\cite{xu2018attngan} & GAN & 256×256 & Image &   \\ \hline

\end{tabular}%
}}
\par
\end{table*}

\subsubsection{Noise-based Generative Methods}
\label{noise-based}
\ 
\newline Noise-based methods utilize random noise as input and map it to the data space through neural network models to generate new data samples. \MM{These methods learn the mapping relationship between noise and data to produce realistic outputs.}

\MM{Recent advancements in generative models have significantly improved the quality and efficiency of both image and video synthesis. Karras et al. \cite{karras2017progressive} pioneered enhancements in image generation by refining generator architecture and training methods, including redesigned normalization, revised progressive growing, and path length regularization. In the realm of video generation, MoCoGAN \cite{tulyakov2018mocogan} innovatively separated motion from content by merging fixed content vectors with stochastic motion vectors, enabling dynamic video generation. Its adversarial learning framework incorporated both image and video discriminators for effective content-motion decomposition. IntroVAE \cite{huang2018introvae} combined the stability of VAEs with the adversarial nature of GANs, training the generator and inference models introspectively to produce high-resolution videos through self-evaluation and refinement. Building on the improvements of \cite{karras2017progressive}, Karras et al. \cite{karras2020analyzing} further refined the approach by incrementally growing the generator and discriminator, accelerating training and supporting high-resolution outputs. Jiang et al. \cite{jiang2021deceive} addressed the challenge of training GANs with limited datasets by introducing Adaptive Pseudo Augmentation (APA), which leverages the generator to expand the real data distribution, preventing discriminator overfitting and enhancing image synthesis quality in data-scarce environments. RV-GAN \cite{gupta2022rv} introduced the TransConv LSTM (TC-LSTM), a novel recurrent GAN architecture that captures spatial and temporal relationships within a single LSTM unit, enabling superior video synthesis. Wang et al. \cite{wang2023styleinv} designed a new motion generator based on a learning-based inversion network, enhancing video generation by constraining the generator’s latent space to incorporate smooth priors and leveraging a pretrained StyleGAN image generator for high-quality video generation with style transfer capabilities.}
\\
\MM{Noise-based generative methods, including GANs and VAEs, are highly flexible in modeling complex data distributions and learning meaningful latent representations, enabling the generation of high-fidelity samples across various domains. However, they face challenges such as mode collapse, training instability, and evaluation difficulties, which can restrict diversity and require meticulous hyperparameter tuning. To overcome these issues, autoregressive models have been developed, leveraging probabilistic dependencies to produce coherent and diverse outputs.} 
\subsubsection{Autoregressive Models}
\label{autoregressive models}
\ 
\newline \MM{Autoregressive models (ARG) formulate the generation task as a conditional probability distribution, progressively generating each element in a sequence based on the previously generated elements, thereby producing a complete data sample. Esser et al. \cite{esser2021imagebart} introduced ImageBART, an innovative ARG model that integrates a multinomial diffusion process and employs a coarse-to-fine context hierarchy to enhance image synthesis. This model excels in high-fidelity image generation and modification, benefiting from efficient training in a compressed latent space, and supports both unconditional and conditional image generation with user-driven local editing. To overcome the limitations of traditional ARG models, such as missing broader scene information, Lee et al. \cite{lee2022autoregressive} proposed a two-stage framework comprising the Residual-Quantized Variational Autoencoder (RQ-VAE) to compress high-resolution images into discrete codes and the RQ-Transformer to model sequence dependencies, enabling rapid generation of high-fidelity images. Huang et al. \cite{huang2023towards} further advanced the field by introducing DynamicQuantization VAE (DQ-VAE) and DQ-Transformer, which encode image regions into variable-length codes based on information density, employing a novel transformer architecture to improve the quality and speed of image generation.}

\MM{The above methods of autoregressive models excel in capturing sequential dependencies and long-range correlations, generating coherent and realistic samples in portrait generation. However, they suffer from slow generation speeds, computational inefficiency, and difficulty in parallelization, and may struggle with complex and multimodal data distributions, limiting their ability to produce diverse, high-fidelity samples. To address these challenges, researchers have turned to Neural Radiance Field models, which offer detailed and realistic representations.}
\subsubsection{Neural Radiance Fields Models}
\label{neural radiance fields models}
\ 
\newline \MM{Neural Radiance Fields (NeRF) are capable of rendering high-quality 3D scene images from any perspective \cite{barron2022mip,wang2021nerf}. In portrait generation, the process begins by producing a NeRF through generative models, which then renders new images by altering the viewpoint to capture previously invisible parts. One of the first attempts at this was GRAF \cite{schwarz2020graf}, which proposed a generative model for radiance fields to achieve precise control over camera viewpoint or object pose. Given a random code and a view direction as NeRF's input, GRAF can synthesize high-resolution images. Niemeyer et al. \cite{niemeyer2021giraffe} advanced this by adopting compositional generative neural feature fields, effectively disentangling individual objects, shapes, and appearances from the background for better manipulation. Bergman et al. \cite{bergman2022generative} introduced the first radiance field for human bodies and facial expressions. They developed a 3D GAN to generate radiance fields representing human bodies and faces in a canonical pose, then warped them through a deformation field. Pix2NeRF \cite{cai2022pix2nerf} was the first work on NeRF-based GAN inversion, leveraging $\pi$-GAN, an unconditional generative model, to transfer random latent codes into object-specific radiance fields.}

\MM{NeRF enhance traditional generative methods by providing detailed and spatially coherent 3D reconstructions, making them particularly effective for generating images from new viewpoints. However, NeRF methods are computationally intensive, have poor generalization to new scenarios, and offer low controllability, making them unsuitable for real-time applications and requiring meticulous tuning for optimal results. To overcome these challenges, researchers have developed conditional generative methods, which introduce specific conditions to guide the generation process, improving both controllability and efficiency.}

\subsection{Conditional Generative Methods}
\label{subsec: conditional generative methods}
\MM{Conditional generative methods enhance unconditional techniques by incorporating specific conditions or parameters to guide image synthesis. It enables the generation of images with predefined attributes, such as labels or textual prompts, allowing precise feature manipulation while maintaining authenticity and resolution. These methods are categorized into label-guided and text-guided synthesis.}

\subsubsection{Label-guided Image/Video Synthesis}
\label{label-guided}
\ 
\newline \MM{Label-guided methods} utilize specific labels or annotations to dictate the characteristics of \MM{generated portraits}. These labels can \MM{encompass} a wide range of attributes such as age, gender, facial expression, or \MM{detailed features like makeup and accessories}. By incorporating these labels into the generative model, the model learns to synthesize portraits that adhere to the specified attributes.
Lu et al. \cite{lu2018attribute} \MM{used} an attribute-guided conditional CycleGAN \MM{to generate} high-resolution face images from low-resolution inputs, controlled by user-defined attributes. Liu et al. \cite{liu2019attribute} \MM{focused on age attributes, presenting a wavelet-based GAN model that captures age-related texture details across multiple scales in frequency space, effectively addressing challenges in face aging with unpaired datasets and mitigating unnatural facial attribute changes.} Men et al. \cite{men2020controllable} \MM{demonstrated the Attribute-Decomposed GAN, which embeds human attributes such as pose and clothing into the latent space as independent codes, enabling flexible control through style-based mixing and interpolation.} Celeb-HQ \cite{zhu2022celebv} is a widely used dataset \MM{for face generation, meticulously annotated with 83 facial attributes, benefiting label-guided techniques significantly.} Zou et al. \cite{zou20234d} \MM{introduced}  a generative framework for creating 3D facial expression sequences using a Denoising Diffusion Probabilistic Model (DDPM) \MM{, which supports both} unconditional and conditional training \MM{with} various inputs such as expression labels and partial sequences.

\MM{Label-guided portrait synthesis offers precise control over generated image attributes, benefiting applications requiring specific visual features and enhancing data efficiency by reducing the need for extensive variant datasets. However, these methods can limit creative freedom and relies heavily on accurately labeled training data, which can introduce biases. To overcome these limitations, researchers have turned to text-guided synthesis, which uses textual descriptions for greater flexibility and control.} 

\subsubsection{Text-guided Image/Video Synthesis}
\label{text-guided}
\ 
\newline Text-guided methods extend the abilities of label-guided techniques by \MM{providing nuanced} and detailed control over generated images through natural language descriptions. \MM{These methods use text descriptions to specify complex combinations of features and scenarios in a more expressive and flexible manner}.
Mansimov et al. \cite{mansimov2015generating} \MM{made the first attempt at image generation from natural language, employing a combination of RNN and VAE, but the results were blurry.} Reed et al. \cite{reed2016generative} \MM{pioneered the use of GANs for synthesizing realistic images from text descriptions, combining advanced recurrent neural networks for text feature extraction with deep convolutional GANs for image generation.} Xu et al. \cite{xu2018attngan} \MM{presented} an attention-driven GAN for detailed text-to-image synthesis\MM{, focusing on relevant words in text descriptions to allow precise control over fine-grained image details.} Canfes et al. \cite{canfes2023text} \MM{enabled control over both shape and texture using text or image-based prompts by manipulating the latent space of 3D generative models with the help of the CLIP model and a pre-trained 3D GAN model. Autoregressive models have also been employed for text-guided image generation using large-scale data.} Ramesh et al. \cite{ramesh2021zero} \MM{introduced DALL-E, a transformer that models text and image tokens as a unified data stream, simplifying the process by treating text and image information equally and performing competitively with sufficient data and scale.} Yu et al. \cite{yu2022scaling} \MM{proposed Parti, which treats the task as a sequence-to-sequence problem similar to machine translation, using a Transformer-based image tokenizer to encode images into discrete tokens. Both Parti and DALL-E suffer from high computation costs due to their large number of parameters.}

\MM{More recently, diffusion models have made major breakthroughs in this task. Huang et al. \cite{huang2023collaborative} detailed Collaborative Diffusion, which enables multi-modal control over the generative process in diffusion models. This approach leverages existing pre-trained uni-modal diffusion models, facilitating multi-modal face generation and editing without the need for re-training.} Wu et al. \cite{wu2024portrait3d} \MM{developed an advanced neural rendering-based approach to text-to-3D-portrait generation with 3DPortraitGAN, producing high-quality 3D portraits consistent with textual descriptions. The quality of text-to-video generation has also improved significantly. Ho et al. \cite{ho2022imagen} presented Imagen Video, an advanced text-conditional video generation system utilizing a cascade of video diffusion models to generate high-definition videos. Wu et al. \cite{wu2023tune} introduced One-Shot Video Tuning for text-guided video generation, which requires only a single text-video pair for training. Tune-A-Video employs a spatiotemporal attention mechanism and an efficient one-shot tuning strategy to capture continuous motion, producing smooth and coherent videos.} 

\MM{Text-guided generation is effective for creating images that match text descriptions, with advancements like Adaptive Variational Parameters (AVP) and Actor-Score Distillation Sampling (ASDS) enhancing 3D portrait generation. However, it faces limitations such as reliance on large pre-trained datasets, biases in training data, high computational resource demands, and poor generalization to unseen text descriptions. Building on the foundation of portrait generation, we now turn to talking head driven mechanisms, focusing on generating realistic talking head videos through various driving methods. }

\section{Driven Mechanisms}
\label{sec: Driven Mechanisms}
\MM{Driven mechanisms are pivotal in advancing the field of talking head synthesis, enabling the creation of realistic and expressive talking head videos. By leveraging various input modalities, these mechanisms control the facial movements and expressions of synthesized characters, ensuring natural and coherent results. The field can be broadly divided into two main approaches: video-driven methods and audio-driven methods.}


\subsection{Video-driven Methods}
\label{subsec:video-driven}
\MM{Video-driven generation aims to create realistic face videos by preserving the identity of a still image while replicating the motion from a driving video. High-resolution video generation requires capturing finer facial details and ensuring temporal consistency, which demands larger datasets and longer training times. Video-driven methods are categorized into two main types: traditional non-learning methods, which use predefined algorithms or manual processes, and advanced learning-based methods, which employ deep learning to generate realistic and expressive facial animations from extensive datasets.}
\subsubsection{Traditional Non-learning Methods}
\label{traditional non-learning methods}
\ 
\newline \MM{Before the advent of deep learning technologies, accomplishing video-driven tasks such as facial animation from video input was a formidable challenge. Early animation techniques \cite{lee1995realistic,chuang2002performance}, including rule-based methods and conventional image processing, made it feasible to track and animate facial features effectively. These traditional methods largely relied on geometric transformations, template-based modifications and hybrid transformations.}
\\
\MM{\textbf{Geometric Transformations} involve using mathematical formulas to manipulate facial features by applying transformations like rotation, scaling, and translation. This approach is straightforward and computationally efficient, allowing for the alteration of facial expressions based on predefined rules or sequences. \textbf{Template-based Modifications} involve adjusting predefined facial templates to match the movements observed in a source video. This method typically uses a standard set of facial templates, such as active appearance models \cite{cootes2001active} and blend shape models \cite{pighin2006synthesizing}, which are modified to fit the actor's facial movements frame by frame \cite{deng2006animating,bouaziz2013online}. \textbf{Hybrid Transformations} combine traditional transformations with deep learning techniques to achieve video-driven talking head synthesis. Thies et al. \cite{thies2016face2face} introduced a real-time facial reenactment method using monocular video streams, where dense photometric consistency measurements track facial expressions across source and target videos, facilitating reenactment through rapid and efficient deformation transfer with meticulous fitting of the inner mouth region. X2Face \cite{wiles2018x2face} employs deep learning techniques but fundamentally relies on geometric transformations to adjust a source face based on a driving face or other input modalities, achieving sophisticated video and image editing.}

\MM{Non-learning methods face significant limitations due to their lack of adaptability and scalability, requiring extensive manual setup and fine-tuning for each new face or expression. They struggle to capture the complexity of human facial expressions because of their reliance on rigid templates or simplistic geometric rules. In contrast, learning-based methods automate facial animation and improve accuracy by learning from vast amounts of data, capturing nuances with greater fidelity.}

\begin{table*}[t]
\centering
\caption{Summary of advanced learning-based methods of video-driven branch, where the Unsupervised, End-to-end, N-shot, means the input needs one image (one-shot) or more images (few-shot), 3D Model, Driven Region, refers to the region can be manipulated by driving.}

\label{table: video}
\scalebox{0.8}{
\resizebox{\textwidth}{!}{%
\begin{tabular}{c|cc|ccc}
\hline

\multicolumn{1}{c|}{} & \multicolumn{2}{c|}{Model} & \multicolumn{3}{c}{Characteristic} \\ \hhline{|>{\arrayrulecolor{white}}->{\arrayrulecolor{black}}|-----|}

\multirow{-2}{*}{\centering{Method}} &
\multicolumn{1}{c|}{Unsupervised} &
\multicolumn{1}{c|}{End-to-end} &
\multicolumn{1}{c|}{N-shot} &
\multicolumn{1}{c|}{3D Model} &
\multicolumn{1}{c}{Driven Region}
\\ \hline
\multicolumn{6}{c}{\textbf{Explicit Modeling Paradigm}} \\ \hline

 FOMM.\cite{FOMM2019} & \multicolumn{1}{c|}{\checkmark} &
  \checkmark & \multicolumn{1}{c|}{One-shot} & \multicolumn{1}{c|}{2D} &
  Whole head  \\ \hline
  TPSM.\cite{TPSM2022}         & \multicolumn{1}{c|}{\checkmark} & \checkmark & \multicolumn{1}{c|}{One-shot} & \multicolumn{1}{c|}{2D} & Whole head  \\ \hline
  DaGAN.\cite{DAGAN2022}         & \multicolumn{1}{c|}{\checkmark} &   & \multicolumn{1}{c|}{One-shot} & \multicolumn{1}{c|}{2D} & Whole head  \\ \hline
  Face-vid2vid.\cite{face-vid2vid2021} & \multicolumn{1}{c|}{\checkmark} &   & \multicolumn{1}{c|}{One-shot} & \multicolumn{1}{c|}{3D} & Whole head  \\ \hline
  Free-headGAN.\cite{free-headGAN2023} & \multicolumn{1}{c|}{}  & \checkmark & \multicolumn{1}{c|}{Few-shot} & \multicolumn{1}{c|}{3D} & Whole head \\ \hline
  Dual-GFR.\cite{hsu2022dual}     & \multicolumn{1}{c|}{}  &   & \multicolumn{1}{c|}{One-shot} & \multicolumn{1}{c|}{3D} & Face     \\ \hline
  MarioNETte.\cite{MarioNETte2019}   & \multicolumn{1}{c|}{\checkmark} &   & \multicolumn{1}{c|}{Few-shot} & \multicolumn{1}{c|}{3D} & Face       \\ \hline
  Metaportrait.\cite{zhang2023metaportrait}  & \multicolumn{1}{c|}{}  &   & \multicolumn{1}{c|}{One-shot} & \multicolumn{1}{c|}{2D} & Whole head \\ \hline
  GONHA.\cite{li2024generalizable}        & \multicolumn{1}{c|}{}  &   & \multicolumn{1}{c|}{One-shot} & \multicolumn{1}{c|}{3D} & Whole head  \\ \hline
 MeshG.\cite{yao2020mesh}    & \multicolumn{1}{c|}{\checkmark} & \checkmark & \multicolumn{1}{c|}{One-shot} & \multicolumn{1}{c|}{3D} & Face    \\ \hline
 ROME.\cite{khakhulin2022ROME}         & \multicolumn{1}{c|}{\checkmark} & \checkmark & \multicolumn{1}{c|}{One-shot} & \multicolumn{1}{c|}{3D} & Whole head \\ \hline
  SAFA.\cite{wang2021safa}         & \multicolumn{1}{c|}{\checkmark} & \checkmark & \multicolumn{1}{c|}{One-shot} & \multicolumn{1}{c|}{3D} & Face       \\ \hline
  Head3D.\cite{Head3D2024}       & \multicolumn{1}{c|}{\checkmark} &   & \multicolumn{1}{c|}{One-shot} & \multicolumn{1}{c|}{3D} & Whole head  \\ \hline

 \multicolumn{6}{c}{\textbf{Implicit Modeling Paradigm}} \\ \hline
  Image Set.\cite{mallya2022implicit}    & \multicolumn{1}{c|}{\checkmark} & \checkmark & \multicolumn{1}{c|}{Few-shot} & \multicolumn{1}{c|}{2D} & Whole head  \\ \hline 
  TS-Net.\cite{TS-Net2023}       & \multicolumn{1}{c|}{\checkmark} &   & \multicolumn{1}{c|}{Few-shot}     & \multicolumn{1}{c|}{2D} & Whole head  \\ \hline
  Fast-bi.\cite{fast-bi2020}      & \multicolumn{1}{c|}{}  & \checkmark & \multicolumn{1}{c|}{One-shot} & \multicolumn{1}{c|}{2D} & Whole head  \\ \hline
  LIA.\cite{LIA2022}          & \multicolumn{1}{c|}{\checkmark} &   & \multicolumn{1}{c|}{One-shot} & \multicolumn{1}{c|}{2D} & Whole head  \\ \hline
   LSR.\cite{meshry2021learned}          & \multicolumn{1}{c|}{}  & \checkmark & \multicolumn{1}{c|}{Few-shot} & \multicolumn{1}{c|}{2D} & Whole head  \\ \hline
  ReenactGAN.\cite{Reenactgan2018}   & \multicolumn{1}{c|}{}  &   & \multicolumn{1}{c|}{One-shot} & \multicolumn{1}{c|}{2D} & Whole head \\ \hline
  LNFR-GAN.\cite{bounareli2022finding}     & \multicolumn{1}{c|}{\checkmark} &   & \multicolumn{1}{c|}{One-shot} & \multicolumn{1}{c|}{3D} & Face      \\ \hline 
  Exp-S2LS.\cite{oorloff2023expressive}   & \multicolumn{1}{c|}{}  & \checkmark & \multicolumn{1}{c|}{One-shot} & \multicolumn{1}{c|}{3D} & Face      \\ \hline
  Hyperreenact.\cite{bounareli2023hyperreenact} & \multicolumn{1}{c|}{}  & \checkmark & \multicolumn{1}{c|}{One-shot} & \multicolumn{1}{c|}{3D} & Face       \\ \hline
  FACEGAN.\cite{tripathy2021facegan}      & \multicolumn{1}{c|}{}  & \checkmark & \multicolumn{1}{c|}{One-shot} & \multicolumn{1}{c|}{3D} & Face       \\ \hline
  HiDeNeRF.\cite{HideNerf2023}     & \multicolumn{1}{c|}{}  &   & \multicolumn{1}{c|}{One-shot} & \multicolumn{1}{c|}{3D} & Whole head \\ \hline
  DNRF.\cite{gafni2021dynamic}         & \multicolumn{1}{c|}{\checkmark} & \checkmark & \multicolumn{1}{c|}{Few-shot} & \multicolumn{1}{c|}{3D} & Whole head \\ \hline
\end{tabular}%
}}
\par
\end{table*}

\subsubsection{Advanced Learning-based Methods}
\label{advanced learning-based methods}
\ 
\newline \MM{Advanced learning-based methods have revolutionized video-driven talking head synthesis by using deep learning to create highly realistic and expressive facial animations. These techniques dynamically learn complex mappings from extensive datasets, significantly enhancing animation quality and accuracy. The approaches are broadly categorized into two paradigms: explicit modeling and implicit modeling.} We summarize advanced learning-based methods of video-driven branch from multiple perspectives in Table \ref{table: video}.
\\
\textbf{Explicit modeling paradigm} \MM{in video-driven talking head synthesis involves the direct manipulation of clearly defined features and parameters, such as keypoints and meshes, making these methods highly interpretable and often simpler to implement. Keypoints-based warping methods generate motion flow by learning the correspondence between keypoints, thereby warping the features of the source image \cite{siarohin2019animating,siarohin2021MRAA}. Siarohin et al. \cite{FOMM2019} developed the First Order Motion Model (FOMM), a method that animates objects in images using keypoints derived via self-supervised learning and first-order local affine transformations, generating a dense motion field and occlusion masks to deform the source image at the encoder's feature layer and restore it through the decoder. Zhao et al. \cite{TPSM2022} improved upon FOMM by predicting multiple sets of keypoints for thin-plate spline transformations and incorporating an affine transformation for background prediction. DAGAN \cite{DAGAN2022} further enhanced FOMM by utilizing depth information to improve the precision of warps and reduce artifacts. Face vid2vid \cite{face-vid2vid2021} introduced dynamic 3D keypoints for unsupervised learning from single-source images, enabling one-shot free-view avatar synthesis. Zhang et al. \cite{zhang2023metaportrait} leveraged dense landmarks to generate precise geometry-aware flow fields, ensuring the adaptive fusion of source identity during synthesis to maintain essential features. Doukas et al. \cite{free-headGAN2023} used sparse 3D facial landmarks for face modeling, eliminating the need for complex facial priors, and incorporated explicit gaze control and an attention mechanism within the generator to facilitate few-shot learning and accommodate multiple source images. To address challenges like inadvertent transfer of the driver's identity and handling unfamiliar large poses, Ha et al. \cite{MarioNETte2019} implemented image attention blocks, target feature alignment, and landmark transformers, with the landmark transformer playing a critical role in preserving identity by decoupling landmarks to isolate expression geometry from identity traits.}

\MM{In contrast to keypoints-based methods, mesh-based rendering methods depend on 3D head reconstruction models and 3D representation models to provide a more comprehensive understanding of human facial dynamics. These models offer a detailed perspective for reenacting facial movements \cite{chen2020talking}. Besides the widely used 3DMM, models like DECA \cite{feng2021learning}, Emoca \cite{danvevcek2022emoca} are also employed. Yao et al. \cite{yao2020mesh} proposed a method that uses 3D mesh reconstruction to guide optical flow learning for facial reenactment. Wang et al. \cite{wang2021safa} utilized 3D Morphable Models (3DMM) for facial simulation while applying multiple affine transformations to render foreground elements like hair and beard. Khakhulin et al. \cite{khakhulin2022ROME} introduced ROME, a mesh-based method for creating one-shot human head avatars from a single photo, determining the head mesh and neural textures from the photograph. Li et al. \cite{li2024generalizable} presented an innovative approach for reconstructing and animating 3D head avatars from single portrait images, using a tri-plane method \cite{chan2022efficient} to capture aspects of 3D geometry, appearance details, and facial expressions.}

\MM{Keypoints-based warping methods maintain subject identity but struggle with occlusion and significant motion, while mesh-based rendering offers higher fidelity but faces challenges like model complexity and high computational demands. These limitations lead to the implicit modeling paradigm, which uses latent embeddings to infer features and relationships, offering greater flexibility and handling complex, non-linear interactions in dynamic facial animations.}
\\
\textbf{Implicit Modeling Paradigm} \MM{in video-driven talking head synthesis utilizes latent embeddings or spaces to infer features and relationships, providing greater flexibility and handling complex, non-linear interactions. Latent space-based methods represent images as embedding codes. Wu et al. \cite{Reenactgan2018} introduced ReenactGAN, the first end-to-end learning-based facial reenactment framework, which translates the source face into a boundary latent space, adjusts it with a transformer, and reconstructs the target frame using a target-specific decoder. Wang et al. \cite{LIA2022} proposed the LIA model, achieving animation by learning orthogonal motion directions within the latent space and linearly combining them, avoiding complex processing based on structural representation. Meshry et al. \cite{meshry2021learned} developed a few-shot face synthesis method that decomposes the target subject's representation into spatial and style components, predicting the latent space layout and using it for spatial denormalization to synthesize the target frame. Zakharov et al. \cite{fast-bi2020} presented a facial rendering technique based on neural networks, hierarchically modeling facial features within an embedding space by processing low-frequency information and high-frequency texture separately. Ni et al. \cite{TS-Net2023} introduced a dual-branch network architecture that captures motion from driving videos and creates high-quality video content. Bounareli et al. \cite{bounareli2022finding} addressed the separation of identity and posture by combining a pre-trained GAN with a 3D shape model, manipulating facial orientations and expressions within the GAN's latent space. Oorloff et al. \cite{oorloff2023expressive} utilized StyleGAN2 image inversion and multi-stage nonlinear latent space editing to reproduce high-resolution facial videos. Bounareli et al. \cite{bounareli2023hyperreenact} further refined this approach by employing a hypernetwork to fine-tune source identity features, addressing visual artifacts and reducing fine-tuning costs.}

\MM{NeRF methods also play a crucial role in video-driven tasks. Gafni et al. \cite{gafni2021dynamic} demonstrated a dynamic NeRF for simulating facial appearance and movement, combining scene representation networks with a low-dimensional deformable model for explicit control over posture and expressions, using volumetric rendering techniques. Hong et al. \cite{hong2022headnerf} presented HeadNeRF, a parametric head model using NeRFs to render high-fidelity human head images, incorporating 2D neural rendering and improving detail accuracy. Li et al. \cite{HideNerf2023} introduced HiDe-NeRF, which decomposes 3D dynamic scenes into canonical appearance and implicit deformation fields, maintaining facial shape and details through multi-scale volumetric features. Other implicit methods include Tripathy et al. \cite{tripathy2021facegan}, who conveyed facial actions through Action Unit (AU) representation to prevent identity information leakage. Mallya et al. \cite{mallya2022implicit} employed an implicit warping approach for motion transfer, using a cross-modal attention layer to generate image animation by creating correspondences between source and driving images. Hong et al. \cite{hong2023implicit} used implicit identity representation to better preserve the source image's identity. However, this lack of explicit structure can reduce model interpretability, complicating fine-tuning and debugging.}

\MM{Video-driven methods have advanced the generation of realistic talking head animations by preserving the identity of a still image while replicating the motion from a driving video. However, they face challenges such as occlusion, maintaining identity integrity, and handling large pose variations. In contrast, audio-driven methods leverage audio signals to synchronize lip movements and facial expressions with spoken content, thereby enhancing the realism and naturalness of talking head animations.}

\subsection{Audio-driven Methods}
\label{subsec:audio-driven}
\MM{Audio-driven talking head synthesis transforms a static facial image into a dynamic video synchronized with audio inputs. This multimodal task faces several challenges due to the inherent differences between audio and visual modalities, such as lip synchronization \cite{lu2021live} and the diversity and authenticity of expressions \cite{eskimez2021speech}. To address these challenges, existing approaches are classified into two main categories: statistical parametric models and cutting-edge deep generative models.} 
We also systematically summarize audio-drien methods in Table \ref{table: audio}.

\begin{table*}[t]
\centering
\caption{Summary of audio-driven methods, where the Unsupervised, End-to-end, N-shot, means the input needs one image (one-shot) or more images (few-shot), 3D Model, Driven Region, refers to the region can be manipulated by driving.}
\label{table: audio}
\scalebox{0.8}{
\resizebox{\textwidth}{!}{%
\begin{tabular}{c|cc|ccc}
\hline
\multicolumn{1}{c|}{} & \multicolumn{2}{c|}{Model} & \multicolumn{3}{c}{Characteristic} \\ \hhline{|>{\arrayrulecolor{white}}-|>{\arrayrulecolor{black}}-----|} 

  \multirow{-2}{*}{\centering{Method}} &
  \multicolumn{1}{c|}{Unsupervised} &
  \multicolumn{1}{c|}{End-to-end} &
  \multicolumn{1}{c|}{N-shot} &
  \multicolumn{1}{c|}{3D Model} &
  \multicolumn{1}{c}{Driven Region} \\ \hline

\multicolumn{6}{c}{\textbf{Statistical Parametric Models} } \\ \hline 
  CHMM.\cite{xie2007coupled}   & \multicolumn{1}{c|}{}                &            & \multicolumn{1}{c|}{N/A}      & \multicolumn{1}{c|}{2D}       & Mouth      \\ \hline
VL-DNN.\cite{zhang2013new}                  & \multicolumn{1}{c|}{}                &            & \multicolumn{1}{c|}{N/A}      & \multicolumn{1}{c|}{2D}       & Mouth       \\ \hline
VOCA.\cite{cudeiro2019capture}                    & \multicolumn{1}{c|}{}                & \checkmark           & \multicolumn{1}{c|}{N/A}      & \multicolumn{1}{c|}{3D}       & Whole head   \\ \hline
DiNET.\cite{zhang2023dinet}                   & \multicolumn{1}{c|}{}                &            & \multicolumn{1}{c|}{One-shot} & \multicolumn{1}{c|}{2D}       & Face       \\ \hline
You said that.\cite{chung2017you}   & \multicolumn{1}{c|}{\checkmark }               &            & \multicolumn{1}{c|}{Few-shot} & \multicolumn{1}{c|}{2D}       & Face      \\ \hline
 DAVD-Net.\cite{zhang2020davd}                & \multicolumn{1}{c|}{}                & \checkmark           & \multicolumn{1}{c|}{Few-shot} & \multicolumn{1}{c|}{2D}       & Face         \\ \hline
Synthesizing Obama.\cite{2017obama}      & \multicolumn{1}{c|}{}                &            & \multicolumn{1}{c|}{N/A}      & \multicolumn{1}{c|}{2D}       & Mouth        \\ \hline
PR-BLSTM.\cite{fan2015photo}     & \multicolumn{1}{c|}{}                &            & \multicolumn{1}{c|}{N/A}      & \multicolumn{1}{c|}{2D}       & Face        \\ \hline
Everybody's token.\cite{song2022everybody}       & \multicolumn{1}{c|}{}                & \checkmark          & \multicolumn{1}{c|}{N/A}      & \multicolumn{1}{c|}{3D}       & Whole head  \\ \hline
MakeItTalk.\cite{zhou2020makelttalk}             & \multicolumn{1}{c|}{}                & \checkmark           & \multicolumn{1}{c|}{One-shot} & \multicolumn{1}{c|}{2D}       & Whole head   \\ \hline
 NVP.\cite{thies2020neural}                     & \multicolumn{1}{c|}{}                & \checkmark           & \multicolumn{1}{c|}{N/A}      & \multicolumn{1}{c|}{3D}       & Whole head   \\ \hline

 \multicolumn{6}{c}{\textbf{Cutting-edge Deep Generative Models} } \\ \hline
FACIAL.\cite{zhang2021facial}                  & \multicolumn{1}{c|}{}                & \checkmark           & \multicolumn{1}{c|}{Few-shot} & \multicolumn{1}{c|}{2D}       & Whole head  \\ \hline
 ATVG.\cite{chen2019hierarchicalATVG}                  & \multicolumn{1}{c|}{}                & \checkmark           & \multicolumn{1}{c|}{One-shot} & \multicolumn{1}{c|}{2D}       & Face        \\ \hline
 MusicFace.\cite{liu2024musicface}               & \multicolumn{1}{c|}{}                &            & \multicolumn{1}{c|}{N/A}      & \multicolumn{1}{c|}{3D}       & Whole head  \\ \hline
RS-DFA.\cite{vougioukas2020realistic}                  & \multicolumn{1}{c|}{}                & \checkmark           & \multicolumn{1}{c|}{One-shot} & \multicolumn{1}{c|}{2D}       & Face        \\ \hline
 DAVS.\cite{zhou2019talking}                    & \multicolumn{1}{c|}{}                & \checkmark           & \multicolumn{1}{c|}{Few-shot} & \multicolumn{1}{c|}{2D}       & Face        \\ \hline
EMMN.\cite{tan2023emmn}                    & \multicolumn{1}{c|}{}                &            & \multicolumn{1}{c|}{Few-shot} & \multicolumn{1}{c|}{2D}       & Face       \\ \hline
 Wav2Lip.\cite{prajwal2020wav2lip}           & \multicolumn{1}{c|}{}                &            & \multicolumn{1}{c|}{Few-shot} & \multicolumn{1}{c|}{2D}       & Face         \\ \hline
 
 FONT.\cite{liu2023font}                    & \multicolumn{1}{c|}{\checkmark }               &            & \multicolumn{1}{c|}{One-shot} & \multicolumn{1}{c|}{2D}       & Whole head   \\ \hline
 DAR.\cite{mittal2020animating}                     & \multicolumn{1}{c|}{}                & \checkmark           & \multicolumn{1}{c|}{N/A}      & \multicolumn{1}{c|}{2D}       & Face        \\ \hline
AG-Codec.\cite{richard2021audio}                & \multicolumn{1}{c|}{}                &            & \multicolumn{1}{c|}{Few-shot} & \multicolumn{1}{c|}{2D}       & Face         \\ \hline
SadTalker.\cite{zhang2023sadtalker}               & \multicolumn{1}{c|}{\checkmark }               & \checkmark           & \multicolumn{1}{c|}{One-shot} & \multicolumn{1}{c|}{2D}       & Whole head   \\ \hline 

 Two-Stage.\cite{bernardo2024speech}               & \multicolumn{1}{c|}{}                &            & \multicolumn{1}{c|}{N/A}      & \multicolumn{1}{c|}{2D}       & Whole head   \\ \hline
FaceFormer.\cite{fan2022faceformer}              & \multicolumn{1}{c|}{\checkmark }               & \checkmark           & \multicolumn{1}{c|}{N/A}      & \multicolumn{1}{c|}{3D}       & Face       \\ \hline
 LipFormer.\cite{wang2023lipformer}               & \multicolumn{1}{c|}{}                & \checkmark           & \multicolumn{1}{c|}{Few-shot} & \multicolumn{1}{c|}{2D}       & Face        \\ \hline

 AD-NeRF.\cite{guo2021ad}                 & \multicolumn{1}{c|}{} &            & \multicolumn{1}{c|}{Few-shot} & \multicolumn{1}{c|}{3D}       & Whole head   \\ \hline
DFA-NeRF.\cite{yao2022dfa}                & \multicolumn{1}{c|}{\checkmark }               &            & \multicolumn{1}{c|}{Few-shot} & \multicolumn{1}{c|}{3D}       & Whole head   \\ \hline 
ER-NeRF.\cite{li2023efficient}                 & \multicolumn{1}{c|}{}                &            & \multicolumn{1}{c|}{N/A}      & \multicolumn{1}{c|}{3D}       & Whole head   \\ \hline 
 GeneFace.\cite{ye2023geneface}                & \multicolumn{1}{c|}{} &            & \multicolumn{1}{c|}{Few-shot} & \multicolumn{1}{c|}{3D}       & Whole head   \\ \hline
DFRF.\cite{shen2022learning}                    & \multicolumn{1}{c|}{}                &            & \multicolumn{1}{c|}{Few-shot} & \multicolumn{1}{c|}{3D}       & Whole head   \\ \hline
SyncTalk.\cite{peng2024synctalk}                  & \multicolumn{1}{c|}{\checkmark }               &   \multicolumn{1}{c|}{\checkmark }         & \multicolumn{1}{c|}{Few-shot} & \multicolumn{1}{c|}{3D}       & Whole head   \\ \hline
  DiffTalk.\cite{shen2023difftalk}   & \multicolumn{1}{c|}{}                &            & \multicolumn{1}{c|}{Few-shot}      & \multicolumn{1}{c|}{2D}       & Whole head       \\ \hline
Diffused Heads.\cite{stypulkowski2024diffused}          & \multicolumn{1}{c|}{}                &            & \multicolumn{1}{c|}{One-shot} & \multicolumn{1}{c|}{2D}       & Whole head  \\ \hline
V-Express.\cite{wang2024v}          & \multicolumn{1}{c|}{}                &            & \multicolumn{1}{c|}{Few-shot} & \multicolumn{1}{c|}{2D}       & Whole head  \\ \hline
Dreamtalk.\cite{ma2023dreamtalk}               & \multicolumn{1}{c|}{\checkmark }               &            & \multicolumn{1}{c|}{One-shot} & \multicolumn{1}{c|}{3D}       & Whole head   \\ \hline
 TGDM.\cite{xu2023multimodal}                    & \multicolumn{1}{c|}{}                &            & \multicolumn{1}{c|}{One-shot} & \multicolumn{1}{c|}{3D}       & Whole head   \\ \hline
ACDM.\cite{bigioi2024speech}                    & \multicolumn{1}{c|}{}                & \checkmark           & \multicolumn{1}{c|}{One-shot} & \multicolumn{1}{c|}{2D}       & Whole head  \\ \hline
PAV-DP.\cite{yu2023talking}                  & \multicolumn{1}{c|}{}                &            & \multicolumn{1}{c|}{One-shot} & \multicolumn{1}{c|}{2D}       & Whole head  \\ \hline

\end{tabular}%
}}
\par
\end{table*}

\subsubsection{Statistical Parametric Models}
\label{statistical parametric models}
\ 
\newline \MM{The evolution of statistical parametric models for audio-driven talking head synthesis can be traced through the development and application of Hidden Markov Models (HMMs), Convolutional Neural Networks (CNNs), and Recurrent Neural Networks (RNNs).}
\\
\MM{\textbf{HMMs-based} facial animation methods effectively capture the dynamics in video and audio sequences \cite{choi1999baum,choi2001hidden}. Yamamoto et al. \cite{yamamoto1998lip} pioneered the use of HMMs in audio-driven animation by developing a method to synthesize lip movements from voice input, effectively addressing coarticulation challenges in lip movement prediction. Subsequent methods based on HMMs evolved to improve upon these initial efforts \cite{lee2002audio,aleksic2004speech}. However, due to the lack of a one-to-one correspondence between phonemes. Xie et al. \cite{xie2007coupled} introduced the coupled hidden Markov model (CHMM), a multi-stream approach that enhances animation realism by modeling the asynchrony and temporal dependencies of audiovisual speech, converting acoustic speech into high-quality facial animation parameters using an Expectation Maximization (EM) algorithm.}
\\
\MM{\textbf{CNNs-based} facial animation methods utilize deep learning architectures to enhance the quality and realism of facial animations. Chung et al. \cite{chung2017you} pioneered the use of CNNs to synthesize animated talking face videos from a single image and its corresponding audio track, presenting an encoder-decoder framework that merges facial and audio data embeddings. With the development of deep learning, Karras et al. \cite{karras2017audio} proposed a DCNN method that incorporates deformable convolutions for reconstructing talking head videos at extremely low bitrates, embedding encoder information from video compression standards and using a constrained projection module to repair compression artifacts, improving reconstruction quality.Cudeiro et al. \cite{cudeiro2019capture} introduced VOCA, a neural network that generates realistic 3D facial animations from speech signals across languages, incorporating diverse speaking styles and identity-specific characteristics without retraining. Zhang et al. \cite{zhang2023dinet} introduced DiNet, a method for few-shot learning in high-resolution facial dubbing, using spatial deformation of feature maps to retain texture details and synchronize lip movements with audio.}
\\
\MM{\textbf{RNN-based} facial animation methods further refine the ability to handle sequential data, with Long Short-Term Memory (LSTM) architectures being the most frequently employed variant \cite{fan2016deep,eskimez2018generating,thies2020neural}. Fan et al. \cite{fan2015photo} used deep Bidirectional Long Short-Term Memory (BLSTM) networks for audio-visual modeling, converting data into parallel time series and training the BLSTM to minimize errors between context labels and visual features for more detailed animation. Suwajanakorn et al. \cite{2017obama} trained a neural network on audio clips of President Obama to map audio features to mouth shapes, synthesizing high-quality lip textures into target videos using 3D pose matching. Song et al. \cite{song2022everybody} proposed a method that decomposes video frames into expression, geometry, and pose parameters, using a recurrent network to transform audio into synchronized mouth movements while preserving geometric and postural context. Zhou et al. \cite{zhou2020makelttalk} introduced a method for generating expressive talking head videos from a single facial image and audio input by separating lip movements and other dynamics, and predicting facial landmarks to reflect the speaker's intentions.}

\MM{The above methods have significantly advanced audio-driven facial animation by enhancing realism and handling sequential data. However, they face limitations in maintaining lip synchronization, achieving diverse and authentic expressions, and personalizing the speaker's style. Cutting-edge deep generative models leverage sophisticated neural architectures and extensive datasets to produce superior results in audio-driven talking head synthesis, addressing these challenges effectively.}

\subsubsection{Cutting-edge Deep Generative Models}
\label{Cutting-edge Deep Generative Models}

\begin{figure*}[!ht]
    \centering
    \includegraphics[width=1.0\textwidth]{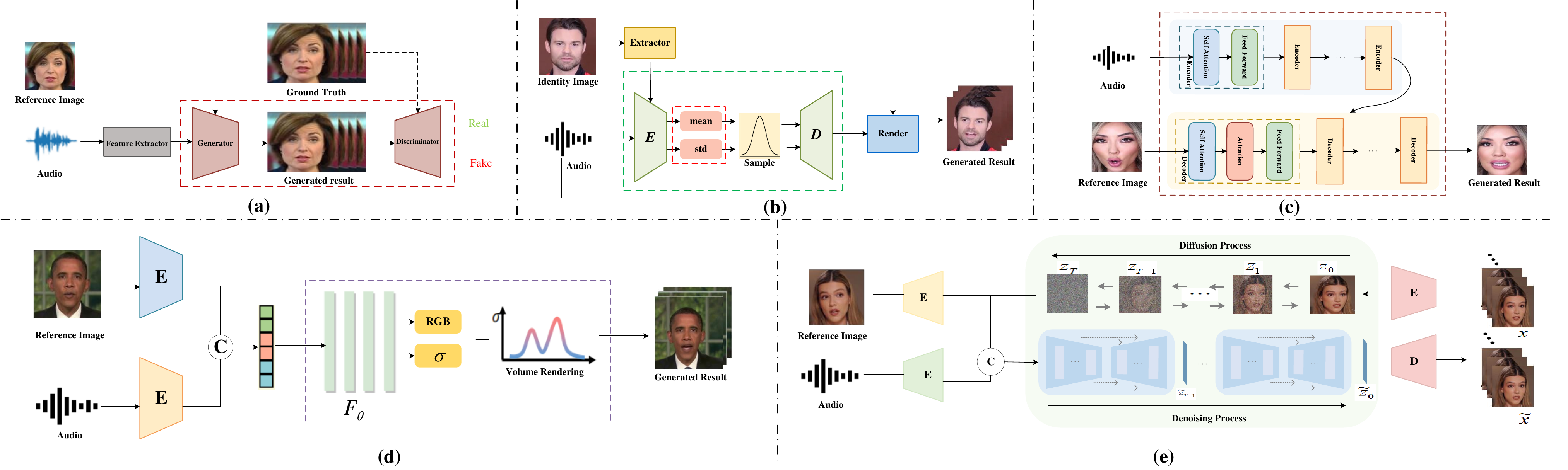}
    \caption{Typical structures for cutting-edge deep generation models of audio-driven branch. (a) GAN-based audiovisual synthesis methods. (b) VAE-based audiovisual mapping methods. (c) Transformer-based sequence translation. (d) Implicit field animation modeling. (e) Diffusive-based expression synthesis methods.  }
    \label{fig: audio-driven} 
\end{figure*}

\ 
\newline \MM{The advent of cutting-edge deep generative models has revolutionized audio-driven talking head synthesis by leveraging advanced neural architectures and extensive datasets to enhance lip synchronization, achieve diverse expressions, and personalize the speaker's style. Key approaches include GAN-based, VAE-based, transformer-based, implicit field-based, explicit parametric-based, and diffusive-based, each introducing unique innovations that advance realistic and expressive talking head animations.}
Figure \ref{fig: audio-driven} compares the main structure of each key approach. 
\\
\MM{\textbf{GAN-based} audiovisual synthesis methods harness the capabilities of GANs to seamlessly integrate auditory cues with visual data \cite{vougioukas2020realistic,zhang2021facial}, enabling precise and realistic facial animations. Chen et al. \cite{chen2019hierarchicalATVG} developed a cascaded GAN that converts audio into facial landmarks before generating video frames, using attention and sequence-aware discriminator to remove pixel jittering and avoid spurious correlations. Zhou et al. \cite{zhou2019talking} developed a method to generate adaptable speaking faces for arbitrary subjects by decoupling subject-specific and speech-related information using joint adversarial training. Prajwal et al. \cite{prajwal2020wav2lip} introduced Wav2Lip, which enhances lip synchronization across various identities in dynamic contexts and establishes rigorous benchmarks for future research, including emotional and music-driven synthesis. Tan et al. \cite{tan2023emmn} propose EMMN, synthesizing speaking facial expressions by combining emotional embeddings and lip movements, ensuring consistent facial expression synthesis. Liu et al. \cite{liu2024musicface} address music-driven singing facial synthesis with a method for natural facial movements and a SingingFace dataset for research. These methods offer substantial improvements in realism and adaptability for talking head synthesis. While they excel in producing high-quality animations and maintaining lip synchronization, challenges such as avoiding spurious correlations and ensuring expression consistency persist.} 
\\
\MM{\textbf{VAE-based} audiovisual mapping methods excel in capturing the complex interdependencies between audio and visual modalities, generating lifelike and expressive facial animations. Mittal et al. \cite{mittal2020animating} introduce an audio representation learning framework that segregates speech content, emotional tone, and background noise, resulting in more precise lip movements even amidst disruptive elements. Liu et al. \cite{liu2023font} propose a flow-guided one-shot model that generates talking head videos with natural head movements by integrating head pose prediction based on a CVAE probabilistic model, unsupervised keypoint extraction, and flow-guided occlusion-aware video synthesis techniques, effectively bridging the audio-visual modality gap. Richard et al. \cite{richard2021audio} demonstrate a real-time 3D facial model that accurately predicts facial coefficients representing geometric and texture features using latent cues from lossy input signals like audio and eye-tracking, based on a multimodal VAE. Zhang et al. \cite{zhang2023sadtalker} introduce SadTalker, which synthesizes 3D motion coefficients from audio inputs and integrates them with advanced 3D perception rendering technology to accurately map audio-motion correlations, capturing detailed facial expressions and head movements. However, their reliance on high-quality training data can lead to overfitting or poor generalization if the data is insufficient or biased. Additionally, their stochastic nature can result in inconsistent animations over longer sequences.}
\\
\MM{\textbf{Transformer-based} sequence translation methods leverage the capabilities of the Transformer model, initially proposed and widely adopted in the field of Natural Language Processing (NLP) to address long-distance dependency issues and improve sequence-to-sequence tasks \cite{han2021transformer,acheampong2021transformer,antoun2020arabert}. Fan et al. \cite{fan2022faceformer} leveraged a transformer-based autoregressive model with a tailored biased attention mechanism to capture long-term audio context, aligning audio-motion modalities, and accurately forecasting animated 3D facial mesh sequences for enhanced lip movement precision. Huang et al. \cite{3DFormer} used a transformer-based encoder to map audio signals to 3DMM parameters, guiding the generation of high-quality avatars by predicting long-term audio context. Wang et al. \cite{wang2023lipformer} presented a method for synthesizing high-fidelity talking heads by using a pre-trained facial image codebook within the Transformer framework, predicting sequences of lip codes from audio features to accurately mimic speech-related lip variations. Bernardo et al. \cite{bernardo2024speech} introduced a dual-stage framework combining transformers to extract facial landmarks from audio and GANs to generate realistic video frames, allowing for dynamic shapes and multiple appearances by adjusting weights. However, these methods require substantial computational resources and extensive datasets to achieve optimal performance. }
\\
\MM{\textbf{Implicit field-based} animation modeling methods utilize advanced 3D reconstruction models to generate highly realistic talking head animations. A prominent approach in this category is NeRF, initially proposed by Mildenhall et al. \cite{mildenhall2021nerf}, which is particularly relevant to this task. Guo et al. \cite{guo2021ad} used neural scene representation networks to generate dynamic neural radiance fields from audio signals, synthesizing high-fidelity talking head videos with flexible adjustments to audio, viewing angles, and backgrounds, and extending to upper body generation. Yao et al. \cite{yao2022dfa} employed a neural radiance field to decouple lip motion from personalized features, predicting lip movements from audio and sampling head movements and blinking through a VAE. Shen et al. \cite{shen2022learning} introduced the Dynamic Facial Radiance Field (DFRF), which uses 2D appearance images and a differentiable facial deformation module to quickly adapt to new identities and generate high-quality talking heads with minimal training data. Li et al. \cite{li2023efficient} adopted conditional neural radiance fields with NeRF-based tri-plane hash representations and spatial regions to generate talking portraits, achieving rapid convergence, real-time rendering, and high efficiency. Ye et al. \cite{ye2023geneface} proposed a method for generating realistic video portraits from arbitrary audio, using a VAE and a conditioned NeRF renderer to address head-body disconnection, enhancing video fidelity and generalizability. These methods offer flexibility in adjusting audio signals, viewing angles, and backgrounds, thereby enhancing the realism and adaptability of the generated animations\cite{peng2024synctalk}. However, they often require extensive computational resources and sophisticated models to achieve optimal results.}  
\\
\MM{\textbf{Explicit parametric-based} modeling methods leverage the strengths of explicit representations to address the limitations of implicit radiance field models. The latest technique, 3D Gaussian, demonstrates superior performance in 3D reconstruction compared to NeRF \cite{zhou2024headstudio}. First, we introduce some research on 3D human head avatar modeling. Xu et al. \cite{xu2023gaussian} utilized controllable 3D Gaussian models for high-fidelity head avatar modeling. By incorporating high-frequency dynamic details through a fully learned MLP-based deformation field, it effectively simulates a wide range of extreme expressions. Zhao et al. \cite{zhao2024psavatar} demonstrated that PSAvatar uses a Point-based Morphable Shape Model (PMSM) with 3D Gaussian modeling, excels in real-time animation by enabling flexible and detailed 3D geometry modeling. Cho et al. \cite{cho2024gaussiantalker} proposed GaussianTalker, a real-time pose-controllable talking head model that integrates 3D Gaussian attributes with audio features into a shared implicit feature space, leveraging 3D Gaussian Splatting (3DGS) for rapid rendering, resulting in enhanced facial fidelity, lip-sync accuracy, and superior rendering speed compared to existing models. However, the parametric nature of 3D Gaussians imposes limitations on the complexity of shapes that can be modeled without a substantial increase in the number of Gaussians, thereby escalating computational costs and potentially affecting efficiency. }
\\
\MM{\textbf{Diffusive-based} expression synthesis methods leverage diffusion models to incrementally add and remove noise from data, transforming it to and from a Gaussian distribution to generate high-quality images \cite{saharia2022palette,gu2022vector}. Yu et al. \cite{yu2023talking} achieved realistic and natural lip-sync by sampling facial movements that are unrelated to lip movements, leveraging the probabilistic characteristics of audio-to-visual diffusion priors to produce varied facial movement sequences for the same audio input. Shen et al. \cite{shen2023difftalk} used latent diffusion models to generate high-resolution, audio-synchronized talking head videos that adapt to new identities without fine-tuning. Ma et al. \cite{ma2023dreamtalk} presented a diffusion model framework that predicts target expressions from audio and generates realistic, style-varied talking heads with precise lip movements. Xu et al. \cite{xu2023multimodal} introduced the Texture-Geometry-aware Diffusion Model (TGDM) that utilizes the CLIP model for nuanced emotional control, employing a multi-conditional denoising task to produce highly realistic talking faces. Stypulkowski et al. \cite{stypulkowski2024diffused} introduced a solution for talking-face generation using autoregressive diffusion models, generating realistic talking head videos from a single identity image and audio sequence. Bigioi et al. \cite{bigioi2024speech} demonstrated an end-to-end audio-driven video editing method using denoising diffusion models to generate synchronized facial movements directly from audio Mel spectrogram features, eliminating the need for intermediate representations like facial landmarks or 3D models. These advancements underscore the efficacy of diffusion-based approaches in producing lifelike, expressive talking heads, setting a new benchmark in audiovisual synthesis.}
\section{Editing Techniques}
\label{sec:editing techniques}
\MM{Editing techniques in the context of talking head synthesis build upon the foundational phases of portrait generation and talking head animation. These techniques are essential for refining and customizing the final outputs to meet specific needs and enhance realism. Editing techniques can be broadly classified into two categories: 2D image/video editing and 3D model editing.} We summarize common methods of editing in Table \ref{table: editing}.

\begin{table*}[t]
\centering
\caption{Summary of editing techniques in talking head synthesis, where the Driven Content, Whole Frame Manipulation, means the method pays attention to the whole image's generation, including person and background, Controllable Attribute, including Lip Sync, Head Pose, Expression, Eye Blink.}
\label{table: editing}
\scalebox{0.8}{
\resizebox{\textwidth}{!}{%
\begin{tabular}{c|c|c|cccc}
\hline

\multicolumn{1}{c|}{} & \multicolumn{1}{c|}{} & \multicolumn{1}{c|}{}&
  \multicolumn{4}{c}{Controllable Attribute} \\ \hhline{|>{\arrayrulecolor{white}}-|>{\arrayrulecolor{white}}-|>{\arrayrulecolor{white}}-|>{\arrayrulecolor{black}}----|}
\multirow{-2}{*}{\centering{Method}} &
  \multirow{-2}{*}{\centering{Driven Content}} &
  \multirow{-2}{*}{\centering{Whole-frame Manipulation}} &
  \multicolumn{1}{c|}{Lip Sync} &
  \multicolumn{1}{c|}{Head Pose} &
  \multicolumn{1}{c|}{Expression} &
   \multicolumn{1}{c}{Eye Blink} \\ \hline

\multicolumn{7}{c}{\textbf{2D Image/Video Editing}} \\ \hline
PECHead.\cite{gao2023PECFace}    & Video            & \checkmark & \multicolumn{1}{c|}{}  & \multicolumn{1}{c|}{\checkmark} & \multicolumn{1}{c|}{\checkmark}          &   \\ \hline
 DPE.\cite{pang2023dpe}        & Video            & \checkmark & \multicolumn{1}{c|}{}  & \multicolumn{1}{c|}{\checkmark} & \multicolumn{1}{c|}{\checkmark}          & \checkmark \\ \hline
 PDRL.\cite{wang2023progressive}       & Audio and image  & \checkmark & \multicolumn{1}{c|}{\checkmark} & \multicolumn{1}{c|}{\checkmark} & \multicolumn{1}{c|}{\checkmark}          & \checkmark \\ \hline
 OPT.\cite{liu2023opt}        & Audio and image  & \checkmark & \multicolumn{1}{c|}{\checkmark} & \multicolumn{1}{c|}{\checkmark} & \multicolumn{1}{c|}{\checkmark}          &   \\ \hline
 GC-AVT.\cite{liang2022expressive}     & Audio and image  &   & \multicolumn{1}{c|}{\checkmark} & \multicolumn{1}{c|}{\checkmark} & \multicolumn{1}{c|}{\checkmark}          &   \\ \hline
ICface.\cite{tripathy2020icface}     & Image            & \checkmark & \multicolumn{1}{c|}{}  & \multicolumn{1}{c|}{\checkmark} & \multicolumn{1}{c|}{\checkmark}          & \checkmark \\ \hline
 PC-AVS.\cite{zhou2021pose}     & Audio and video  &   & \multicolumn{1}{c|}{\checkmark} & \multicolumn{1}{c|}{\checkmark} & \multicolumn{1}{c|}{\checkmark}          &   \\ \hline

\multicolumn{7}{c}{\textbf{3D Model Editing}} \\ \hline
HeadGAN.\cite{doukas2021headgan}    & Video            & \checkmark & \multicolumn{1}{c|}{}  & \multicolumn{1}{c|}{\checkmark} & \multicolumn{1}{c|}{\checkmark}          &   \\ \hline
 OTAvatar.\cite{ma2023otavatar}   & Video           &   & \multicolumn{1}{c|}{}  & \multicolumn{1}{c|}{\checkmark} & \multicolumn{1}{c|}{\checkmark}          &   \\ \hline
 PIRenderer.\cite{ren2021pirenderer} & Video or audio   & \checkmark & \multicolumn{1}{c|}{\checkmark} & \multicolumn{1}{c|}{\checkmark} & \multicolumn{1}{c|}{\checkmark}          &   \\ \hline
 3DTF-PPD.\cite{zhang20213d}   & Audio            & \checkmark & \multicolumn{1}{c|}{\checkmark} & \multicolumn{1}{c|}{\checkmark} & \multicolumn{1}{c|}{}           &   \\ \hline
 NED.\cite{papantoniou2022neural}        & Video            & \checkmark & \multicolumn{1}{c|}{}  & \multicolumn{1}{c|}{\checkmark} & \multicolumn{1}{c|}{\checkmark}          & \checkmark \\ \hline
 Styletalk.\cite{ma2023styletalk}  & Audio and video  & \checkmark & \multicolumn{1}{c|}{\checkmark} & \multicolumn{1}{c|}{\checkmark} & \multicolumn{1}{c|}{\checkmark}          &   \\ \hline
 TalkCLIP.\cite{ma2023talkclip}   & Audio and text   & \checkmark & \multicolumn{1}{c|}{\checkmark} & \multicolumn{1}{c|}{\checkmark} & \multicolumn{1}{c|}{\checkmark}          &   \\ \hline
 TBEHead.\cite{fried2019text}    & Text             & \checkmark & \multicolumn{1}{c|}{\checkmark} & \multicolumn{1}{c|}{\checkmark} & \multicolumn{1}{c|}{}           & \checkmark \\ \hline

\end{tabular}%
}}
\par
\end{table*}

\subsection{2D Image/Video Editing}
\label{subsec: 2D editing}
\MM{2D image/video editing for talking heads involves advanced techniques that enhance and modify facial animations by manipulating specific visual attributes while maintaining overall video coherence, divided into two primary approaches: disentanglement and decoupling, and control and modulation.}
\subsubsection{Disentanglement and Decoupling}
\label{Disentanglement and Decoupling}
\ 
\newline \MM{These techniques focus on separating and independently controlling various aspects of facial animations in talking heads. By isolating different attributes such as expression, identity, and pose, these methods enable precise and flexible editing without affecting other visual components. Liang et al. \cite{liang2022expressive} present a method for granular control of a speaking head by decomposing the reference driving image into a cropped mouth, a masked head, and the upper face, achieving a blend of lip synchronization, pose control, and emotional expression. Pang et al. \cite{pang2023dpe} presented a self-supervised decoupling framework that operates independently of 3D Morphable Models (3DMM) and paired data. This framework employs a motion editing module to decouple and independently control pose and expression in the latent space, making it suitable for video editing. Wang et al. \cite{wang2023progressive} introduced a method for fine-grained control over lip movement, blink, head pose, and emotional expression by progressively decoupling motion features within unstructured video data, effectively leveraging and separating these motion factors. Tan et al. \cite{tan2024edtalk} introduced EDTalk, which accommodates diverse input modalities and flexible control by dividing facial dynamics into three orthogonal latent spaces representing the mouth, pose, and expression, ensuring their independence and allowing for more refined disentanglement and application.}

\MM{The disentanglement and decoupling techniques have significantly advanced the precision and flexibility of editing talking head animations by isolating and independently controlling various facial attributes such as expression, identity, and pose. These methods offer the advantage of targeted manipulation without compromising other visual components, resulting in highly detailed and customizable animations. However, the complexity of decoupling multiple attributes can introduce challenges in maintaining naturalness and coherence, particularly when dealing with diverse and dynamic input data. }

\subsubsection{Control and Modulation}
\label{Control and Modulation}
\ 
\newline \MM{Technologies in 2D image and video editing focus on fine-tuning and precisely manipulating specific aspects of facial animations. These methods aim to achieve nuanced control over various facial features, such as lip movements, eye blinks, and emotional expressions, enabling the creation of highly realistic and expressive talking heads. Tripathy et al. \cite{tripathy2020icface} proposed a facial animator that controls pose and expressions using human-interpretable signals like head pose angles and action unit values, supporting mixed control sources and selective post-editing. Liu et al. \cite{liu2023opt} designed a network for generating pose-controllable talking heads, using an audio feature disentanglement module to enhance facial structure and identity preservation while enabling free pose control. Zhou et al. \cite{zhou2021pose} presented a framework for pose-controllable talking faces, modularizing audio-visual representations with an implicit low-dimension pose code, effectively operating on non-aligned raw face images. Sun et al. \cite{sun2024fg} advanced fine-grained controllable facial expression editing by using input action units (AUs) to control muscle group intensities and integrating AU features with the expression latent code.}

\MM{Control and modulation techniques in 2D image and video editing have shown significant advancements in achieving detailed and precise facial animations. These methods offer flexibility in manipulating specific facial features and maintaining the realism of talking heads. However, they often require complex models and extensive data to ensure accuracy and naturalness. }

\subsection{3D Model Editing}
\label{subsec:3D editing}
\MM{3D model editing techniques represent the forefront of advancements in talking head synthesis, offering unparalleled precision and realism by directly manipulating three-dimensional facial representations. Unlike 2D methods, which often face limitations in capturing complex facial geometries and depth, 3D model editing provides a robust framework for achieving lifelike animations. The field is categorized into two primary approaches: direct parametric control and Multimodal integration.}

\subsubsection{Direct Parametric Control}
\label{Direct Parametric Control}
\ 
\newline \MM{These techniques focus on manipulating predefined parameters to achieve precise adjustments in 3D facial models. By leveraging parametric models, they offer fine-grained control over specific facial attributes such as expressions, head poses, and skin textures. Ren et al. \cite{ren2021pirenderer} introduced a model utilizing 3DMM parameters for controlling facial movements from a single image, adaptable for audio-driven facial reenactment by extracting actions from audio input. Doukas et al. \cite{doukas2021headgan} designed a HeadGAN, which edits expressions and poses by adapting 3D facial representations from any driving video to the facial geometry of a reference image, effectively decoupling identity and expressions. Zhang et al. \cite{zhang20213d} presented a method that uses PoseGAN to generate head poses from audio input and PGFace to create a natural facial model based on these poses. Gao et al. \cite{gao2023PECFace} proposed a high-fidelity talking head synthesis method with freely controllable head poses and expressions, combining self-supervised and 3D model-based landmarks with a motion-aware multi-scale feature alignment module to effectively convey motion information. Ma et al. \cite{ma2023otavatar} presented OTAvatar, which uses a controllable tri-plane rendering approach to quickly create personalized talking head avatars from a single reference portrait. It converts the portrait into an identity code and merges it with a motion code, effectively separating identity from motion for efficient CNN-based generation and volumetric rendering.}

\MM{Direct parametric control techniques can provide precise adjustments in 3D facial models. However, the reliance on predefined parameters can limit flexibility and the ability to capture more complex, dynamic interactions in facial expressions. This approach often requires extensive manual tuning and can struggle with generalizing to unseen data.}

\subsubsection{Multimodal Integration}
\label{Multimodal Integration}
\ 
\newline \MM{This type involves extracting control information from multimodal inputs and using it to guide the parameterization or adjustment of 3DMM. By leveraging diverse data sources, such as audio, text, and visual cues, these methods create dynamic and contextually appropriate facial animations. Fried et al. \cite{fried2019text} developed a method for editing talking head videos by modifying the transcription text to alter video content while maintaining audio-visual fluency. It automatically annotates videos with key parameters, enabling tasks such as adding, deleting, and modifying words, language conversion, and sentence creation. Ma et al. \cite{ma2023talkclip} presented a one-shot method for generating talking heads controlled by natural language descriptions. By leveraging a text-video dataset and a CLIP-based style encoder, it aligns textual descriptions with speaking styles, generating realistic talking heads without the need for reference videos. Papantoniou et al. \cite{papantoniou2022neural} proposed a method for altering an actor's emotional state in a video using only an emotional label to control facial expressions while maintaining lip movements, utilizing 3D facial representation and a deep domain transformation framework. Ma et al. \cite{ma2023styletalk} designed a one-shot style-controllable talking face generation framework that extracts speaking style from a reference video and applies it to a single portrait with audio, generating talking head videos with diversified styles.}

\MM{Multimodal Integration techniques offer significant advantages in generating contextually appropriate and dynamic facial animations. However, these methods will be limited by the quality and variety of the input data and the complexity of effectively integrating multimodal information, which can impact the consistency and realism of the generated animations.}

\section{Benchmarking: Datasets, Evaluation Metrics, and Experimental Analysis}
\label{sec:benchmarking}


\subsection{Datasets}
\label{subsec:datasets}
\MM{With the rapid advancements in deep learning, the field of talking head synthesis has seen significant progress, largely due to the availability of open-source datasets that provide essential data support. We have compiled and summarized detailed information on commonly used and recently released datasets (including three announced in 2022) in Table \ref{table: dataset}, along with their access links.}

\begin{table*}[t]
\caption{Detailed information of datasets in talking head synthesis.}
\label{table: dataset}
\resizebox{\textwidth}{!}{%
\begin{tabular}{c|c|c|c|c|c|c|c|c|c|c}
\hline
Dataset  & Year & Hour & Speaker & Sentence & Head Movement & Image Size & Environment & URL & Modality    & View         \\ \hline
GRID.\cite{GRID}    & 2006 & 27.5  & 33  & 33k  & \texttimes  & 360\texttimes288, 720\texttimes576    & Lab   & \href{https://www.grid.ac/}{Link}  & Video, audio   & Frontal view \\ \hline
CREMA-D.\cite{CREMA-D}   & 2014 & 11.1  & 91 & 12    & \checkmark    & 960\texttimes720             & Lab   & \href{https://github.com/CheyneyComputerScience/CREMA-D}{Link}  & Video, audio  &  N/A  \\ \hline
TCD-TIMIT.\cite{TCD}   & 2015 & 11.1  & 62  & 6.9k   & \texttimes  & 1920\texttimes1080                & Lab    & \href{https://sigmedia.github.io/resources/dataset/tcd_timit/}{Link}    & Video, audio  & Multi-view   \\ \hline
LRW.\cite{LRW}             & 2016 & 173   & 1k+ & 539k  & \texttimes  & 256\texttimes256            & Wild        &  \href{https://www.robots.ox.ac.uk/~vgg/data/lip_reading/lrw1.html}{Link}     & Video, text   & Frontal-view \\ \hline
MSP-IMMPROV.\cite{MSP}     & 2016 & 18 & 12  & 652  & \checkmark      & 1440\texttimes1080  & Lab   & \href{https://ecs.utdallas.edu/research/researchlabs/msp-lab/MSP-Improv.html}{Link}  & Audio, video, image   & Frontal-view \\ \hline
Voxceleb.\cite{VOX1}        & 2017 & 352   & 1.2k  & 153.5k & \checkmark             & 224\texttimes224       & Wild      & \href{https://www.robots.ox.ac.uk/~vgg/data/voxceleb/}{Link}  & Video         &  N/A  \\ \hline
ObamaSet.\cite{2017obama}       & 2017 & 14    & 1 & - & \checkmark  & N/A     & Wild        & \href{https://github.com/supasorn/synthesizing_obama_network_training}{Link}    & Video    &   N/A  \\ \hline
MODALITY.\cite{MODALITY}    & 2017 & 31    & 35      & 5.8k     & \texttimes      & 1920\texttimes1080   & Lab         & \href{http://www.modality-corpus.org/}{Link}     & Audio, video    & N/A \\ \hline
Voxceleb2.\cite{vox2}       & 2018 & 2.4k  & 6.1k  & 1.1m  & \checkmark             & 224\texttimes224   & Wild     & \href{https://www.robots.ox.ac.uk/~vgg/data/voxceleb/vox2.html}{Link}     & Video     & N/A \\ \hline
LRS2-BBC.\cite{LRSBBC}        & 2018 & 224.5 & 500+  & 140k+  & \checkmark             & 224\texttimes224     & Wild        & \href{https://www.robots.ox.ac.uk/~vgg/data/lip_reading/lrs2.html}{Link}     & Video, audio, text   & Multi-view \\ \hline
LRS3-TED.\cite{lrs3}        & 2018 & 438   & 5k+   & 152k+ & \checkmark            & 224\texttimes224 & Wild        & \href{https://www.robots.ox.ac.uk/~vgg/data/lip_reading/lrs3.html}{Link}     & Video, audio, text      & Multi-view   \\ \hline
Lombard.\cite{alghamdi2018corpus} & 2018 & 3.6 & 54 & 5.4k & \checkmark & \begin{tabular}[c]{@{}c@{}}720\texttimes480, 864\texttimes480\end{tabular} & Lab & \href{https://spandh.dcs.shef.ac.uk//avlombard/}{Link}  & Video, audio & Multi-view \\ \hline
MELD.\cite{poria2018meld}    & 2018 & 13.7  & 407     & 13.7k    & \checkmark             &    N/A      & Wild        & \href{https://affective-meld.github.io/}{Link}     & Video, audio, text    &   N/A  \\ \hline
RAVDESS.\cite{livingstone2018ryerson}    & 2018 & 7     & 24      & 2   & \checkmark             & 1920\texttimes1080, 1280\texttimes720 & Lab      & \href{https://www.kaggle.com/datasets/uwrfkaggler/ravdess-emotional-speech-audio}{Link}     & Video, audio & Frontal-view \\ \hline
LRW-1000.\cite{yang2019lrw}    & 2019 & 57    & 2k+     & 718k     & \checkmark             & 1024\texttimes576, 1920\texttimes1080 & Wild        & \href{https://paperswithcode.com/dataset/lrw-1000}{Link}     & Video, audio, text         & Frontal-view \\ \hline
Faceforensics++.\cite{rossler2019faceforensics++} & 2019 & 5.7   & 1k    & 1k+   & \checkmark             & 512\texttimes512            & Wild        & \href{https://github.com/ondyari/FaceForensics}{Link}     & Video, image    & Frontal-view \\ \hline
FFHQ.\cite{karras2019style}            & 2019 & -   & -  & -    & \texttimes   & 1024\texttimes1024             & Wild  & \href{https://github.com/NVlabs/ffhq-dataset}{Link}     & Image    &  N/A  \\ \hline
MEAD.\cite{wang2020mead}       & 2020 & 39    & 60   & 20  & \checkmark   & 1920×1080   & Lab    & \href{https://github.com/uniBruce/Mead}{Link}     & Video, audio              & Multi-view   \\ \hline
HDTF.\cite{zhang2021flow}  & 2021 & 15.8  & 362  & 10k & \checkmark   & 1280\texttimes720, 1920\texttimes1080       & Wild        & \href{https://github.com/MRzzm/HDTF}{Link}     & Video    &  N/A \\ \hline
TalkingHead-1KH.\cite{wang2021one} & 2021 & 1000  &  -  &  -  & \checkmark    &            N/A              & Wild        & \href{https://github.com/tcwang0509/TalkingHead-1KH}{Link}     & Video                    &       N/A       \\ \hline
CelebV-HQ.\cite{zhu2022celebv}   & 2022 & 65  & 15653   &   -   & \checkmark        & 512\texttimes512-more          & Wild        & \href{https://celebv-hq.github.io/}{Link}     & Video      &   N/A    \\ \hline
Multiface.\cite{wuu2022multiface}       & 2022 &   -    & 13      &  - & \checkmark    & 2048\texttimes1334, 1024\texttimes1024      & Lab         & \href{https://github.com/facebookresearch/multiface}{Link}     & Image, audio              & Multi-view   \\ \hline
\end{tabular}%
}
\end{table*}

\MM{For video-driven tasks, the most frequently adopted datasets include Voxceleb \cite{VOX1}, Voxceleb2 \cite{vox2}, and TalkingHead-1KH \cite{face-vid2vid2021}. For audio-driven tasks, the key datasets are CREMA-D \cite{CREMA-D}, LRW \cite{LRW}, and MEAD \cite{wang2020mead}. As application scenarios expand, the demands for higher image quality, particularly in terms of detail representation and visual effects, have grown. Consequently, several large-scale, high-resolution datasets have emerged in recent years, such as FFHQ \cite{karras2019style}, HDTF \cite{zhang2021flow}, and CelebV-HQ \cite{zhu2022celebv}. More importantly, we have gathered comprehensive information for each dataset, including image size, modality, and the view of the subject, which to our knowledge has not been extensively covered in previous reviews of talking head synthesis.}

\subsection{Evaluation Metrics}
\label{subsec:Evaluation Metrics}
\MM{A comprehensive evaluation of talking head synthesis models is crucial for understanding their effectiveness. Chen et al. \cite{chen2020comprises} suggested evaluating these models based on four key criteria: identity preservation, visual quality, lip synchronization, and natural motion. While these dimensions are critical, models rarely excel in all areas simultaneously, often focusing on advancements in a single aspect. In the domain of talking head synthesis, both qualitative and quantitative assessments are widely used. While qualitative assessments rely on direct observation and can be subjective, quantitative metrics provide a more objective evaluation. This section introduces several widely recognized quantitative metrics that are crucial for a thorough evaluation of talking head synthesis models.}
\\
\textbf{Structural Similarity (SSIM)} \cite{wang2004image} \MM{is a metric that measures the similarity between two images, aligning with the human visual system's perception of image quality. It evaluates the similarity between the final results and the ground truth based on luminance, contrast, and structure.}
\\
\textbf{Peak Signal-to-Noise Ratio (PSNR)} \MM{is a widely used metric for assessing image quality. It is calculated as the ratio of the peak signal energy to the mean square error between the generated frame and the ground truth, providing a measure of the fidelity of the generated image.}
\\
\textbf{Frechet Inception Distance (FID)} \cite{heusel2017gans} \MM{evaluates the quality of generated images by measuring the similarity between the distributions of real and generated images in the feature space. This metric calculates the distance between these two distributions, with features extracted from the Inception v3 network \cite{szegedy2016rethinking}, to provide an accurate reflection of image quality.}
\\
\textbf{Learned Perceptual Image Patch Similarity (LPIPS)} \cite{zhang2018unreasonable} \MM{employs a pre-trained convolutional neural network to extract features from image pairs, weights these features, and then computes the $L_2$ norm of the weighted features to derive an LPIPS score. It is crucial for assessing image quality, as it correlates closely with human visual perception.}
\\
\textbf{Inception Score (IS)} \cite{salimans2016improved} \MM{quantifies both the quality and diversity of generated images. Despite its utility, IS has limitations and is often used in conjunction with FID for a more comprehensive evaluation.}
\\
\textbf{Landmark Distance (LMD)} \cite{chen2018lip} \MM{measures the precision of lip movements in synthesized videos by calculating the Euclidean distance between corresponding landmarks in the lip region, normalized by temporal length and the number of landmarks.}
\\
\textbf{Average Keypoint Distance (AKD)} \MM{measures the average distance between key points in the generated image and the corresponding key points in the ground truth. It is used to assess how well the motion from the input driving image is preserved in the generated output.}
\begin{table*}[t]
\centering
\caption{Comparison of state-of-the-art methods in talking head synthesis which takes an image and a video as inputs. Quantitative evaluations are conducted on various datasets, with the best performance for each metric highlighted in blue.}
\label{table: video quatitative}
\scalebox{0.8}{
\resizebox{\textwidth}{!}{%
\begin{tabular}{c|c|c|ccccc}
\hline
\multirow{2}{*}{Dataset} & \multirow{2}{*}{Method} & \multirow{2}{*}{Editing} & \multicolumn{5}{c}{Metrics} \\ \cline{4-8} 
 &  &  &\multicolumn{1}{c|}{SSIM↑} & \multicolumn{1}{c|}{PSNR↑} & \multicolumn{1}{c|}{LPIPS↓} & \multicolumn{1}{c|}{FID↓} & AKD↓ \\ \hline
\multirow{10}{*}{Voxceleb\cite{VOX1}} & FOMM\cite{FOMM2019} & & \multicolumn{1}{c|}{0.723} & \multicolumn{1}{c|}{30.390} & \multicolumn{1}{c|}{0.199} & \multicolumn{1}{c|}{-} & 1.294 \\ \cline{2-8} 
 & Face-vid2vid\cite{face-vid2vid2021} &  & \multicolumn{1}{c|}{0.761} & \multicolumn{1}{c|}{30.690} & \multicolumn{1}{c|}{0.212} & \multicolumn{1}{c|}{-} & 1.620 \\ \cline{2-8} 
 & Bi-layer\cite{fast-bi2020} &  &\multicolumn{1}{c|}{-} & \multicolumn{1}{c|}{20.190} & \multicolumn{1}{c|}{0.152} & \multicolumn{1}{c|}{92.200} & - \\ \cline{2-8} 
 & DAGAN\cite{DAGAN2022} & & \multicolumn{1}{c|}{0.804} & \multicolumn{1}{c|}{31.220} & \multicolumn{1}{c|}{0.185} & \multicolumn{1}{c|}{-} & 1.279 \\ \cline{2-8} 
 & MCNet\cite{hong2023implicit} &  &\multicolumn{1}{c|}{\textbf{\color{data-1}0.825}} & \multicolumn{1}{c|}{\textbf{\color{data-1}31.94}} & \multicolumn{1}{c|}{0.174} & \multicolumn{1}{c|}{-} &{\textbf{\color{data-1}1.203}} \\ \cline{2-8} 
 & MarioNETte\cite{MarioNETte2019} &  &\multicolumn{1}{c|}{0.744} & \multicolumn{1}{c|}{23.240} & \multicolumn{1}{c|}{-} & \multicolumn{1}{c|}{-} & - \\ \cline{2-8} 
 & HeadGAN\cite{doukas2021headgan} & \checkmark &\multicolumn{1}{c|}{-} & \multicolumn{1}{c|}{21.460} & \multicolumn{1}{c|}{0.112} &  \multicolumn{1}{c|}{36.100} & - \\ \cline{2-8} 
 & Free-HeadGAN\cite{free-headGAN2023} &  &\multicolumn{1}{c|}{-} & \multicolumn{1}{c|}{22.160} & \multicolumn{1}{c|}{\textbf{\color{data-1}0.100}} &  \multicolumn{1}{c|}{35.400} & - \\ \cline{2-8} 
 & HyperReenact\cite{bounareli2023hyperreenact} &  &\multicolumn{1}{c|}{-} & \multicolumn{1}{c|}{-} & \multicolumn{1}{c|}{0.230} & \multicolumn{1}{c|}{\textbf{\color{data-1}27.100}} & - \\ \cline{2-8} 
 & X2Face\cite{wiles2018x2face} &  &\multicolumn{1}{c|}{0.719} & \multicolumn{1}{c|}{22.540} & \multicolumn{1}{c|}{-} & \multicolumn{1}{c|}{-} & 7.687 \\ \hline
\multirow{10}{*}{Voxceleb2\cite{vox2}} & FOMM\cite{FOMM2019} &  &\multicolumn{1}{c|}{0.770} & \multicolumn{1}{c|}{23.250} & \multicolumn{1}{c|}{0.090} & \multicolumn{1}{c|}{73.710} & 2.140 \\ \cline{2-8} 
 & Face-vid2vid.\cite{face-vid2vid2021} &  &\multicolumn{1}{c|}{0.800} & \multicolumn{1}{c|}{24.370} & \multicolumn{1}{c|}{-} & \multicolumn{1}{c|}{69.130} & 2.070 \\ \cline{2-8} 
 & Bi-layer\cite{fast-bi2020} &  &\multicolumn{1}{c|}{0.660} & \multicolumn{1}{c|}{16.980} & \multicolumn{1}{c|}{\textbf{\color{data-1}0.080}} & \multicolumn{1}{c|}{203.360} & 5.380 \\ \cline{2-8} 
 & PIRender\cite{ren2021pirenderer} & \checkmark &\multicolumn{1}{c|}{-} & \multicolumn{1}{c|}{-} & \multicolumn{1}{c|}{0.264} & \multicolumn{1}{c|}{14.407} & - \\ \cline{2-8} 
 & Metaportrait\cite{zhang2023metaportrait} &  &\multicolumn{1}{c|}{-} & \multicolumn{1}{c|}{-} & \multicolumn{1}{c|}{0.226} & \multicolumn{1}{c|}{\textbf{\color{data-1}11.953}} & - \\ \cline{2-8} 
 & X2Face\cite{wiles2018x2face} &  &\multicolumn{1}{c|}{-} & \multicolumn{1}{c|}{-} & \multicolumn{1}{c|}{0.681} & \multicolumn{1}{c|}{45.291} & - \\ \cline{2-8} 
 & ROME\cite{khakhulin2022ROME} &  &\multicolumn{1}{c|}{\textbf{\color{data-1}0.860}} & \multicolumn{1}{c|}{26.200} & \multicolumn{1}{c|}{\textbf{\color{data-1}0.080}} & \multicolumn{1}{c|}{-} & - \\ \cline{2-8} 
 & LIA\cite{LIA2022} &  &\multicolumn{1}{c|}{-} & \multicolumn{1}{c|}{22.290} & \multicolumn{1}{c|}{-} & \multicolumn{1}{c|}{30.230} & 1.049 \\ \cline{2-8} 
 & DAGAN\cite{DAGAN2022} &  &\multicolumn{1}{c|}{-} & \multicolumn{1}{c|}{25.640} & \multicolumn{1}{c|}{-} & \multicolumn{1}{c|}{24.920} & 0.844 \\ \cline{2-8} 
 & PECHead\cite{gao2023PECFace} & \checkmark &\multicolumn{1}{c|}{-} & \multicolumn{1}{c|}{\textbf{\color{data-1}26.960}} & \multicolumn{1}{c|}{-} & \multicolumn{1}{c|}{23.050} & {\textbf{\color{data-1}0.626}} \\ \hline
\multirow{4}{*}{HDTF\cite{zhang2021flow}} & Metaportrait\cite{zhang2023metaportrait} &  &\multicolumn{1}{c|}{-} & \multicolumn{1}{c|}{-} & \multicolumn{1}{c|}{0.208} & \multicolumn{1}{c|}{\textbf{\color{data-1}21.497}} & - \\ \cline{2-8} 
 & ROME\cite{khakhulin2022ROME} &  &\multicolumn{1}{c|}{0.838} & \multicolumn{1}{c|}{20.750} & \multicolumn{1}{c|}{0.173} & \multicolumn{1}{c|}{31.550} & 2.938 \\ \cline{2-8} 
 & OTAvatar\cite{ma2023otavatar} & \checkmark &\multicolumn{1}{c|}{0.806} & \multicolumn{1}{c|}{20.120} & \multicolumn{1}{c|}{0.198} & \multicolumn{1}{c|}{36.630} & 2.933 \\ \cline{2-8} 
 & GONHA\cite{li2024generalizable} &  &\multicolumn{1}{c|}{\textbf{\color{data-1}0.868}} & \multicolumn{1}{c|}{\textbf{\color{data-1}22.150}} & \multicolumn{1}{c|}{\textbf{\color{data-1}0.117}} & \multicolumn{1}{c|}{21.600} & {\textbf{\color{data-1}2.596}} \\ \hline
\multirow{8}{*}{TalkHead-1KH\cite{wang2021one}} & FOMM.\cite{FOMM2019} &  &\multicolumn{1}{c|}{0.790} & \multicolumn{1}{c|}{23.280} & \multicolumn{1}{c|}{0.160} & \multicolumn{1}{c|}{33.220} & 2.905 \\ \cline{2-8} 
 & Face-vid2vid.\cite{face-vid2vid2021}  &  &\multicolumn{1}{c|}{\textbf{\color{data-1}0.810}} & \multicolumn{1}{c|}{23.590} & \multicolumn{1}{c|}{0.160} & \multicolumn{1}{c|}{35.120} & 3.100 \\ \cline{2-8} 
 & MRAA.\cite{siarohin2021MRAA} &  &\multicolumn{1}{c|}{-} & \multicolumn{1}{c|}{25.500} & \multicolumn{1}{c|}{-} &  \multicolumn{1}{c|}{32.570} & 1.057 \\ \cline{2-8} 
 & TPSM.\cite{TPSM2022} &  &\multicolumn{1}{c|}{-} & \multicolumn{1}{c|}{25.530} & \multicolumn{1}{c|}{-} & \multicolumn{1}{c|}{32.770} & 0.983 \\ \cline{2-8} 
 & LIA\cite{LIA2022} &  &\multicolumn{1}{c|}{-} & \multicolumn{1}{c|}{24.430} & \multicolumn{1}{c|}{-} & \multicolumn{1}{c|}{38.890} & 0.932 \\ \cline{2-8} 
 & DAGAN.\cite{DAGAN2022} &  &\multicolumn{1}{c|}{-} & \multicolumn{1}{c|}{23.930} & \multicolumn{1}{c|}{-} & \multicolumn{1}{c|}{34.350} & 2.405 \\ \cline{2-8} 
 & PECHead\cite{gao2023PECFace} &  &\multicolumn{1}{c|}{-} & \multicolumn{1}{c|}{\textbf{\color{data-1}26.760}} & \multicolumn{1}{c|}{-} & \multicolumn{1}{c|}{\textbf{\color{data-1}30.100}} & {\textbf{\color{data-1}0.746}}\\ \cline{2-8} 
 & IWA.\cite{mallya2022implicit} &  &\multicolumn{1}{c|}{-} & \multicolumn{1}{c|}{23.320} & \multicolumn{1}{c|}{\textbf{\color{data-1}0.150}} & \multicolumn{1}{c|}{-} & 3.480 \\ \hline
\end{tabular}%
}}
\par
\end{table*}

\begin{table*}[t]
\centering
\caption{Comparative overview of advanced methods in talking head synthesis using image and audio inputs. Quantitative assessments have been performed across multiple datasets with the superior results marked in blue.}
\label{table: audio quatitative}
\scalebox{0.8}{
\resizebox{\textwidth}{!}{%
\begin{tabular}{c|c|c|ccccc}
\hline
\multirow{2}{*}{Dataset} & \multirow{2}{*}{Method} & \multirow{2}{*}{Editing} &\multicolumn{5}{c}{Metrics} \\ \cline{4-8} 
 &  &  & \multicolumn{1}{c|}{SSIM↑} & \multicolumn{1}{c|}{PSNR↑} & \multicolumn{1}{c|}{LMD↓} & \multicolumn{1}{c|}{CPBD↑} & SyncNet↑ \\ \hline
\multirow{9}{*}{LRW\cite{LRW}}  & Wav2Lip\cite{prajwal2020wav2lip} & &\multicolumn{1}{c|}{\textbf{\color{data-1}0.862}} & \multicolumn{1}{c|}{30.630} & \multicolumn{1}{c|}{5.250} & \multicolumn{1}{c|}{0.152} & {\textbf{\color{data-1}6.900}} \\ \cline{2-8} 
 & MakeItTalk\cite{zhou2020makelttalk} & & \multicolumn{1}{c|}{0.796} & \multicolumn{1}{c|}{30.380} & \multicolumn{1}{c|}{7.130} & \multicolumn{1}{c|}{0.161} & 3.100\\ \cline{2-8} 
 & ATVG\cite{chen2019hierarchicalATVG} & & \multicolumn{1}{c|}{0.810} & \multicolumn{1}{c|}{30.910} & \multicolumn{1}{c|}{\textbf{\color{data-1}1.370}} & \multicolumn{1}{c|}{\textbf{\color{data-1}0.189}} & - \\ \cline{2-8} 
 & PC-AVS\cite{zhou2021pose} & \checkmark& \multicolumn{1}{c|}{0.778} & \multicolumn{1}{c|}{30.390} & \multicolumn{1}{c|}{3.930} & \multicolumn{1}{c|}{0.185} & 6.400 \\ \cline{2-8} 
 & LPGG\cite{chen2018LPGG} & &\multicolumn{1}{c|}{0.730} & \multicolumn{1}{c|}{29.650} & \multicolumn{1}{c|}{1.730} & \multicolumn{1}{c|}{-} & - \\ \cline{2-8} 
 & AVCT\cite{wang2022AVCT} & & \multicolumn{1}{c|}{0.805} & \multicolumn{1}{c|}{-} & \multicolumn{1}{c|}{3.560} & \multicolumn{1}{c|}{0.181} & - \\ \cline{2-8} 
 & EAMM\cite{ji2022eamm}& & \multicolumn{1}{c|}{0.740} & \multicolumn{1}{c|}{\textbf{\color{data-1}30.920}} & \multicolumn{1}{c|}{1.610} & \multicolumn{1}{c|}{-} & 5.520 \\ \cline{2-8} 
 & FONT\cite{liu2023font} & & \multicolumn{1}{c|}{0.825} & \multicolumn{1}{c|}{-} & \multicolumn{1}{c|}{3.480} & \multicolumn{1}{c|}{0.187} & - \\ \cline{2-8} 
 & You said that\cite{chung2017you}& & \multicolumn{1}{c|}{0.770} & \multicolumn{1}{c|}{29.910} & \multicolumn{1}{c|}{1.630} & \multicolumn{1}{c|}{-} & - \\ \hline
\multirow{8}{*}{CREMA-D\cite{CREMA-D}} & Wav2Lip\cite{prajwal2020wav2lip}& & \multicolumn{1}{c|}{0.886} & \multicolumn{1}{c|}{\textbf{\color{data-1}34.230}} & \multicolumn{1}{c|}{3.320} & \multicolumn{1}{c|}{0.253} & 5.890 \\ \cline{2-8} 
 & Audio2Head\cite{wang2021audio2head}& & \multicolumn{1}{c|}{0.59} & \multicolumn{1}{c|}{29.760} & \multicolumn{1}{c|}{3.280} & \multicolumn{1}{c|}{-} & 2.190 \\ \cline{2-8} 
 & MakeItTalk\cite{zhou2020makelttalk} & & \multicolumn{1}{c|}{0.750} & \multicolumn{1}{c|}{31.370} & \multicolumn{1}{c|}{3.360} & \multicolumn{1}{c|}{0.152} & 3.500 \\ \cline{2-8} 
 & PC-AVS\cite{zhou2021pose} & \checkmark& \multicolumn{1}{c|}{0.610} & \multicolumn{1}{c|}{28.470} & \multicolumn{1}{c|}{3.110} & \multicolumn{1}{c|}{0.127} & {\textbf{\color{data-1}6.120}} \\ \cline{2-8} 
 & ACDM\cite{bigioi2024speech} & & \multicolumn{1}{c|}{\textbf{\color{data-1}0.920}} & \multicolumn{1}{c|}{32.470} & \multicolumn{1}{c|}{-} & \multicolumn{1}{c|}{\textbf{\color{data-1}0.290}} & 4.980 \\ \cline{2-8} 
 & EAMM\cite{ji2022eamm} & &\multicolumn{1}{c|}{0.740} & \multicolumn{1}{c|}{29.430} & \multicolumn{1}{c|}{3.120} & \multicolumn{1}{c|}{0.100} & 3.980 \\ \cline{2-8} 
 & RS-DFA\cite{vougioukas2020realistic} & &\multicolumn{1}{c|}{0.844} & \multicolumn{1}{c|}{27.980} & \multicolumn{1}{c|}{-} & \multicolumn{1}{c|}{0.277} & - \\ \cline{2-8} 
 & EMMN\cite{tan2023emmn} & &\multicolumn{1}{c|}{0.680} & \multicolumn{1}{c|}{30.030} & \multicolumn{1}{c|}{\textbf{\color{data-1}3.030}} & \multicolumn{1}{c|}{-} & 2.410 \\ \hline
\multirow{9}{*}{HDTF\cite{zhang2021flow}} 
& DiNET\cite{zhang2023dinet} & & \multicolumn{1}{c|}{\textbf{\color{data-1}0.943}} & \multicolumn{1}{c|}{\textbf{\color{data-1}30.008}} & \multicolumn{1}{c|}{-} & \multicolumn{1}{c|}{-} & {\textbf{\color{data-1}6.840}} \\ \cline{2-8} 
 & PC-AVS\cite{zhou2021pose} & \checkmark&\multicolumn{1}{c|}{0.6383} & \multicolumn{1}{c|}{20.630} & \multicolumn{1}{c|}{3.180} & \multicolumn{1}{c|}{0.164} & 0.420 \\ \cline{2-8} 
 & Audio2Head\cite{wang2021audio2head} & &\multicolumn{1}{c|}{0.735} & \multicolumn{1}{c|}{-} & \multicolumn{1}{c|}{4.830} & \multicolumn{1}{c|}{0.145} & -\\ \cline{2-8} 
 & Wav2Lip\cite{prajwal2020wav2lip} & & \multicolumn{1}{c|}{0.908} & \multicolumn{1}{c|}{29.288} & \multicolumn{1}{c|}{2.890} & \multicolumn{1}{c|}{0.176} & 0.630 \\ \cline{2-8} 
 & MakeItTalk\cite{zhou2020makelttalk} & &\multicolumn{1}{c|}{0.751} & \multicolumn{1}{c|}{19.860} & \multicolumn{1}{c|}{5.460} & \multicolumn{1}{c|}{0.132} & 0.570 \\ \cline{2-8} 
 & AVCT\cite{wang2022AVCT} & &\multicolumn{1}{c|}{0.769} & \multicolumn{1}{c|}{-} & \multicolumn{1}{c|}{2.710} & \multicolumn{1}{c|}{0.167} & 0.760 \\ \cline{2-8} 
 & StyleHEAT\cite{yin2022styleheat} & & \multicolumn{1}{c|}{0.805} & \multicolumn{1}{c|}{19.180} & \multicolumn{1}{c|}{-} & \multicolumn{1}{c|}{-} & - \\ \cline{2-8} 
 & DreamTalk & &\multicolumn{1}{c|}{0.850} & \multicolumn{1}{c|}{-} & \multicolumn{1}{c|}{\textbf{\color{data-1}2.150}} & \multicolumn{1}{c|}{\textbf{\color{data-1}0.310}} & 5.170 \\ \cline{2-8} 
 & SadTalker\cite{zhang2023sadtalker} & & \multicolumn{1}{c|}{0.770} & \multicolumn{1}{c|}{-} & \multicolumn{1}{c|}{4.070} & \multicolumn{1}{c|}{0.240} & 4.350 \\ \hline
\multirow{11}{*}{MEAD\cite{wang2020mead}} 
& DiNET\cite{zhang2023dinet} & &\multicolumn{1}{c|}{\textbf{\color{data-1}0.920}} & \multicolumn{1}{c|}{29.118} & \multicolumn{1}{c|}{-} & \multicolumn{1}{c|}{-} & {\textbf{\color{data-1}7.260}}\\ \cline{2-8} 
 & PC-AVS\cite{zhou2021pose} & \checkmark &\multicolumn{1}{c|}{0.775} & \multicolumn{1}{c|}{23.695} & \multicolumn{1}{c|}{4.970} & \multicolumn{1}{c|}{0.070} & 2.180 \\ \cline{2-8} 
 & Wav2Lip\cite{prajwal2020wav2lip} & & \multicolumn{1}{c|}{0.899} & \multicolumn{1}{c|}{28.539} & \multicolumn{1}{c|}{4.050} & \multicolumn{1}{c|}{\textbf{\color{data-1}0.180}} & 5.260 \\ \cline{2-8} 
 & GC-AVT\cite{liang2022expressive} & \checkmark&\multicolumn{1}{c|}{0.340} & \multicolumn{1}{c|}{-} & \multicolumn{1}{c|}{7.100} & \multicolumn{1}{c|}{0.140} & 2.420 \\ \cline{2-8} 
 & EAMM\cite{ji2022eamm} & &\multicolumn{1}{c|}{0.400} & \multicolumn{1}{c|}{29.230} & \multicolumn{1}{c|}{6.480} & \multicolumn{1}{c|}{0.080} & 1.410 \\ \cline{2-8} 
 & Styletalk\cite{ma2023styletalk} &\checkmark &\multicolumn{1}{c|}{0.840} & \multicolumn{1}{c|}{-} & \multicolumn{1}{c|}{3.250} & \multicolumn{1}{c|}{0.160} & 3.470 \\ \cline{2-8} 
 & EMMN\cite{tan2023emmn} & & \multicolumn{1}{c|}{0.660} & \multicolumn{1}{c|}{\textbf{\color{data-1}29.380}} & \multicolumn{1}{c|}{\textbf{\color{data-1}2.780}} & \multicolumn{1}{c|}{-} & 3.570 \\ \cline{2-8} 
 & SadTalker\cite{zhang2023sadtalker} & & \multicolumn{1}{c|}{0.690} & \multicolumn{1}{c|}{-} & \multicolumn{1}{c|}{4.370} & \multicolumn{1}{c|}{0.160} & 2.760 \\ \cline{2-8} 
 & PD-FGC\cite{wang2023progressive} & &\multicolumn{1}{c|}{0.490} & \multicolumn{1}{c|}{-} & \multicolumn{1}{c|}{4.100} & \multicolumn{1}{c|}{0.050} & 2.270 \\ \cline{2-8} 
 & DreamTalk\cite{ma2023dreamtalk} & &\multicolumn{1}{c|}{0.860} & \multicolumn{1}{c|}{-} & \multicolumn{1}{c|}{2.910} & \multicolumn{1}{c|}{0.160} & 3.780 \\ \cline{2-8} 
 & Audio2Head\cite{wang2021audio2head} & &\multicolumn{1}{c|}{0.600} & \multicolumn{1}{c|}{29.070} & \multicolumn{1}{c|}{3.220} & \multicolumn{1}{c|}{-} & 3.200 \\ \hline
\end{tabular}%
}}
\par
\end{table*}

\subsection{Experimental Analysis}
\label{subsec:experimental analysis}
We compile quantitative results of some classic and SOTA methods on commonly used datasets for both video-driven task and audio-driven task as shown in Table \ref{table: video quatitative} and Table \ref{table: audio quatitative}.
We select commonly used metrics for comparison, moreover, we invest the different method's performance in the same dataset in order to provide an intuitive reference. The two tables both reveal that the performance of the same method varies across different datasets and there isn't a single method performs best across all metrics.

\section{Applications}
\label{sec:application}
Talking head synthesis technology has a wide range of applications across various sectors, revolutionizing how we interact with digital content. This section explores some of the most promising application areas where talking head synthesis is making significant impacts.
\\
 \textbf{Film and television.} Talking head synthesis is transforming visual storytelling in the film and television industry. By employing digital characters, filmmakers can transcend traditional filming constraints. This technology enables the seamless creation of intricate close-up shots and high-risk action sequences that are challenging or unsafe to capture with live actors. Additionally, it reduces reliance on expensive special effects in post-production, thereby lowering production costs and accelerating the creative process. The integration of talking head synthesis not only enhances the visual appeal and narrative depth of productions but also opens up new possibilities for innovative storytelling techniques.
 \\
\textbf{Virtual anchoring.} Virtual anchors, as distributors of news and entertainment content, can broadcast continuously, providing uniform performances in both live and pre-recorded formats. They overcome language and cultural barriers by presenting content in multiple languages to cater to global media consumption needs. The flexibility of virtual anchors in style and appearance allows them to quickly adapt to changing market trends and audience preferences.
\\
\textbf{Game industry.} Talking head synthesis technology significantly enhances the gaming experience by enabling characters to exhibit a wider range of expressions and reactions, thereby increasing player immersion. It allows developers to create characters with rich emotional depth and diverse interactions, offering a unique and interactive gaming experience. As the technology advances, game characters can increasingly reflect players' choices and emotions in real-time, creating personalized gaming. 
\\
\textbf{Virtual instructors.} Talking head synthesis technology enables the creation of virtual educators capable of real-time interaction and expressive facial responses. These virtual instructors deliver standardized educational content and provide personalized guidance based on students' reactions and progress. This interactive and individualized approach enhances student engagement and learning efficiency, making it particularly beneficial for remote education and self-study contexts. The use of virtual instructors significantly expands the reach and accessibility of educational resources.

\section{Summary and Future Work}
\label{sec: summary and future work}
\MM{In this section, we provide a comprehensive overview of the advantages and limitations of current techniques, as well as potential directions for future research. By examining the strengths and weaknesses of existing methods, we can better understand their impact and identify areas for improvement. Furthermore, we will explore promising avenues for future work that can address current challenges and push the boundaries of what talking head synthesis can achieve.}
\subsection{Advantage and Limitation}
\label{subsec: advantage and limitation}
\MM{Portrait generation has advanced significantly through various generative methods, both unconditional and conditional. Unconditional methods, such as noise-based generative models, enable the random creation of portraits without requiring prior labels or data inputs, resulting in diverse and abundant image datasets essential for training many deep learning models. Conditional methods, including text-guided image synthesis, provide controlled and specific outputs based on descriptive inputs, enhancing the relevance of generated portraits for specific applications. Despite these advancements, achieving consistently high-quality and realistic portraits remains challenging. Due to complex geometry and appearance, existing methods face issues like inflexibility, instability, and low fidelity. While large models like diffusion models offer promise, they demand substantial computational resources.}

\MM{The driving mechanism for animating still facial images, or talking head synthesis, has shown impressive performance with both audio-driven and video-driven methods. Audio-driven methods use vocal or text inputs to animate facial features synchronously, while video-driven methods replicate facial expressions and motions from a driving video. Advanced deep learning techniques, including GANs and VAEs, have significantly improved the real-time rendering capabilities of animated portraits, making them viable for interactive applications such as virtual meetings or customer service avatars. However, persistent challenges such as temporal inconsistency and poor results when there are extreme differences between the source image and the driving image hinder further improvement in generation quality. Additionally, the complexity of current models necessitates deep technical knowledge and extensive training data to achieve optimal results.}

\MM{Editing in talking head synthesis, whether 2D-based or 3D-based, has evolved to offer significant control over the modification of generated avatars. Techniques for disentanglement and decoupling allow for the independent manipulation of specific facial features without affecting the overall structure, enhancing flexibility in post-generation customization. Direct parametric control and multimodal integration enable comprehensive adjustments that are coherent across different modalities. However, decoupling itself is challenging due to the interconnected nature of various attributes, and the effectiveness of decoupling directly influences the quality of the results. Ensuring harmony and consistency among attributes is crucial. While 2D-based methods are limited in fine control due to the lack of geometric information, 3D-based methods excel in this aspect but require extensive training time and computational resources.}

\MM{Recent advancements in portrait generation, driving mechanisms, and editing techniques have been significant, yet persistent challenges remain. The integration and complementarity among these domains have opened new possibilities for creating dynamic and lifelike digital portraits but also highlight areas needing further research and optimization. Section \ref{subsec: future work} addresses pivotal issues and provides insights into potential future directions for tackling these challenges.}

\subsection{Future Work}
\label{subsec: future work}
This section will explore the limitations currently faced and propose directions for future research to overcome these challenges and expand the capabilities of these technologies. We have reviewed a plenty of relative papers and summarized the following issues.
\\
\textbf{Dependency on Pretrained Models.} Most current frameworks that utilize pretrained large models, such as diffusion models, heavily rely on the specific capabilities of these models. This dependency can limit innovation and the ability to adapt to new or evolving use cases. Future research could focus on training each module on large-scale datasets or integrating the strengths of pretrained models with other modules designed for specific tasks, thereby reducing reliance on the capabilities of any single model.
\\
\textbf{Data Bias Problem.} In portrait generation, data bias poses a significant challenge, especially in large-scale datasets. This bias can stem from various factors such as shooting environments, ethnicity, gender, and age, leading to generated images that disproportionately exhibit certain attributes or features. This affects the diversity and realism of the generated portraits. To address this issue, incorporating attribute control mechanisms or editing capabilities within the generation process could enhance the model's adaptability to a broader range of attributes and preferences, thereby mitigating bias and improving overall image quality. 
\\
\textbf{One-shot Image as Input.}
Using a single picture as input often fails to provide sufficient data for generating dynamic avatars. This limitation can result in avatars with unnatural, abrupt transitions in expressions and awkward movements, as the static image cannot fully capture the subtle and complex features of talking. Furthermore, if the training data lacks diversity or the model inadequately learns the variety of facial dynamics, the generated images often appear poor when viewed from different angles. Future research should focus on enhancing the accuracy and simplicity of 3D reconstruction techniques to provide more geometric details for generation, thereby improving the overall quality and realism of the avatars.
\\
\textbf{Challenges of Large-Angle Driving.}
Large-angle poses present significant challenges in talking head synthesis due to the lack of training data featuring such angles, resulting in poor model performance. The facial geometry and textures change dramatically with larger head turns, and certain facial parts may be obstructed or incomplete. Addressing these issues requires expanding training datasets to include more images and videos with large-angle poses and developing models that integrate information from multiple viewpoints for a comprehensive 3D facial reconstruction.
\\
\textbf{Temporal consistency problem.} Temporal consistency is critical in talking head synthesis to ensure smooth and natural results. If models fail to handle time-series information effectively, they may not capture dynamic facial expressions and head movements accurately, leading to jumps, jitters, or other unnatural artifacts in generated videos. The quality of the dataset is also vital; datasets with poor frame continuity and insufficient dynamic information can negatively affect the final output. Therefore, methods for handling time-series information and building high-quality datasets are essential for improving temporal consistency.
\\
\textbf{Jittering Issue.}
Jittering arises primarily from a model's inability to maintain smooth motion between consecutive frames. This can be attributed to several factors, including the quality of the training data, the precision of the model in handling subtle facial movements, and the algorithm's management of changes in lighting and shadows during dynamic reconstruction. To address jittering, solutions include enhancing the model's capacity to manage temporal dependencies, optimizing training strategies, and implementing more precise post-processing techniques. 
\\
\textbf{Multilingual Challenges.}
Audio-driven talking head synthesis systems are often optimized for English due to extensive datasets and high demand. This creates challenges when extending to other languages, such as French \cite{dahmani2019conditional}, German \cite{thies2020neural} and Brazilian Portuguese \cite{bernardo2024speech}, which have unique phonetic, rhythmic, and intonational patterns. These differences can result in less accurate and natural animations. Additionally, many languages lack high-quality annotated datasets. Addressing these issues involves collecting diverse multilingual datasets and using meta-learning and transfer learning to improve model adaptability.
\section{Conclusion}
\label{sec:conclusion}
\MM{This article has systematically reviewed and analyzed the key areas of talking head synthesis, including portrait generation, driving mechanisms, and editing techniques. Each domain was explored in detail, highlighting significant papers and summarizing their strengths and limitations. We also compiled comprehensive datasets and evaluated existing techniques using a diverse set of metrics. Furthermore, we discussed widespread applications and outlined future research directions.}

\bibliographystyle{IEEEtran}
\bibliography{refs}

\end{document}